\documentclass{article}
\usepackage{PRIMEarxiv}
\usepackage{authblk}
\usepackage{times}
\usepackage{soul}
\usepackage{url}
\usepackage{graphicx}
\usepackage{amsmath}
\usepackage{amsthm}
\usepackage{amsfonts}
\usepackage{booktabs}
\usepackage{multirow}
\usepackage{algorithm}
\usepackage{subfigure}
\usepackage{stfloats}
\usepackage{pifont}
\usepackage{makecell}
\usepackage{color}
\usepackage{xcolor}
\usepackage{comment}
\usepackage{tcolorbox}
\usepackage{CJKutf8}
\usepackage{algorithmicx}
\usepackage[noend]{algpseudocode}
\usepackage{cuted}
\usepackage{cite}

\def\eg{{e.g., }}

\def\MODEL{\textsf{FlexiCrime}}

\def\cc{{F}} % crime context
\def\vt{{\boldsymbol{t}}} % vector time
\def\vg{{\boldsymbol{g}}} % vector grid
\def\vs{{\boldsymbol{s}}} % vector grid
\def\vr{{\boldsymbol{v}}} % vector target-related

\newtheorem{example}{Example}

\newcommand{\LeftComment}[1]{\Statex \(\triangleright\) #1}

\title{An Event-centric Framework for Predicting Crime Hotspots with Flexible Time Intervals}

\author[1]{Jiahui Jin\thanks{Corresponding author: jjin@seu.edu.cn}~~}
\author[1]{Yi Hong}
\author[2,3]{Guandong Xu}
\author[1]{Jinghui Zhang}
\author[1]{Jun Tang}
\author[1]{Hancheng Wang}
\affil[1]{The School of Computer Science and Engineering, Southeast University}
\affil[2]{University of Technology Sydney}
\affil[3]{The Education University of Hong Kong}

\begin{document}
\maketitle

\begin{abstract}
Predicting crime hotspots in a city is a complex and critical task with significant societal implications. Numerous spatiotemporal correlations and irregularities pose substantial challenges to this endeavor.
Existing methods commonly employ fixed-time granularities and sequence prediction models. However, determining appropriate time granularities is difficult, leading to inaccurate predictions for specific time windows. For example, users might ask: What are the crime hotspots during 12:00-20:00?
To address this issue, we introduce \MODEL, a novel event-centric framework for predicting crime hotspots with flexible time intervals.  \MODEL~incorporates a continuous-time attention network to capture correlations between crime events, which learns crime context features, representing general crime patterns across time points and locations. Furthermore, we introduce a type-aware spatiotemporal point process that learns crime-evolving features, measuring the risk of specific crime types at a given time and location by considering the frequency of past crime events. The crime context and evolving features together allow us to predict whether an urban area is a crime hotspot given a future time interval. To evaluate \MODEL's effectiveness, we conducted experiments using real-world datasets from two cities, covering twelve crime types. The results show that our model outperforms baseline techniques in predicting crime hotspots over flexible time intervals. 
\end{abstract}

\section{Introduction}
Crime prediction is a crucial and challenging task with significant societal implications \cite{zhao2018crime, XuC05}. Accurately predicting crime hotspots can assist law enforcement agencies in efficiently allocating their limited resources to prevent crime and ensure public safety \cite{wortley2016environmental}. However, crime events of various types exhibit strong spatiotemporal correlations and are characterized by extreme sparsity and irregularity \cite{MikeMaguire20177CD}, making the development of effective prediction models challenging \cite{HsinchunChen2004CrimeDM}. Exsiting works \cite{LianghaoXia2021SpatialTemporalSH, li2022spatial,10.1145/3511808.3557657,LianghaoXia2021SpatialTemporalSH,ChaoHuang2018DeepCrimeAH,10508389,Zhao_Liu_Cheng_Zhao_2022} often divide time into fixed intervals and utilize sequence prediction models to forecast crime activity hotspots. Despite their successes, these methods have strict requirements, necessitating that the time interval for prediction aligns with the interval used during training \cite{ChaoHuang2018DeepCrimeAH, XianWu2020HierarchicallyST}. The following example illustrate the challenges associated with solutions based on fixed time intervals.

\begin{figure}[t]
  \centering
  \subfigure[True crime hotspots during various time intervals]{
    \begin{minipage}[t]{0.65\linewidth}
      \centering
      \includegraphics[width=\linewidth]
      {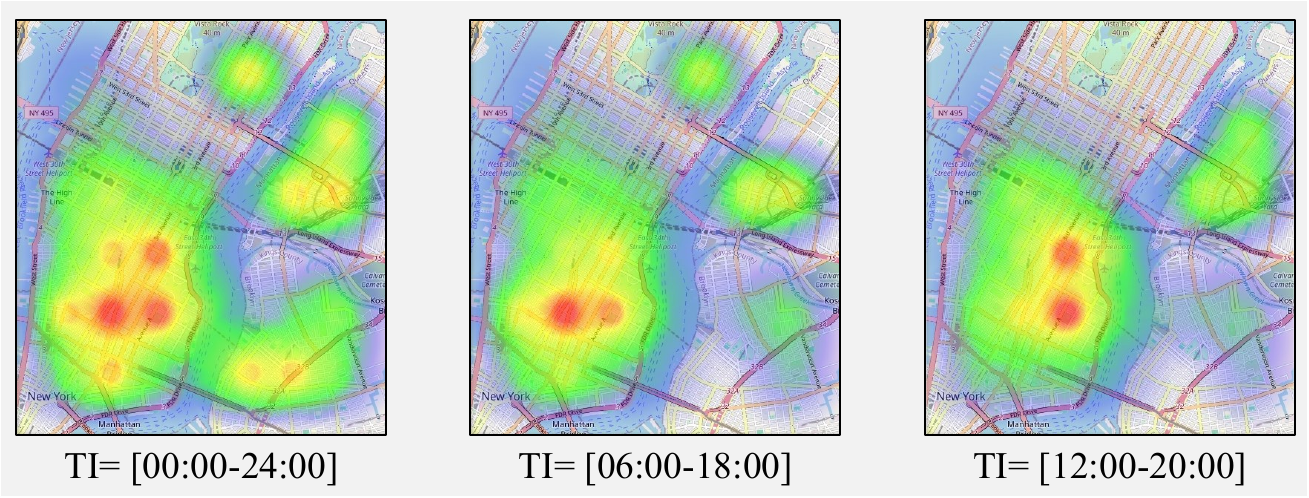}
      \label{fig:intro1}
    \end{minipage}
  }
  \\
  \subfigure[Prediction results for  fixed time intervals with specified time granularities]{
    \begin{minipage}[t]{0.65\linewidth}
      \centering
      \includegraphics[width=\linewidth]
      {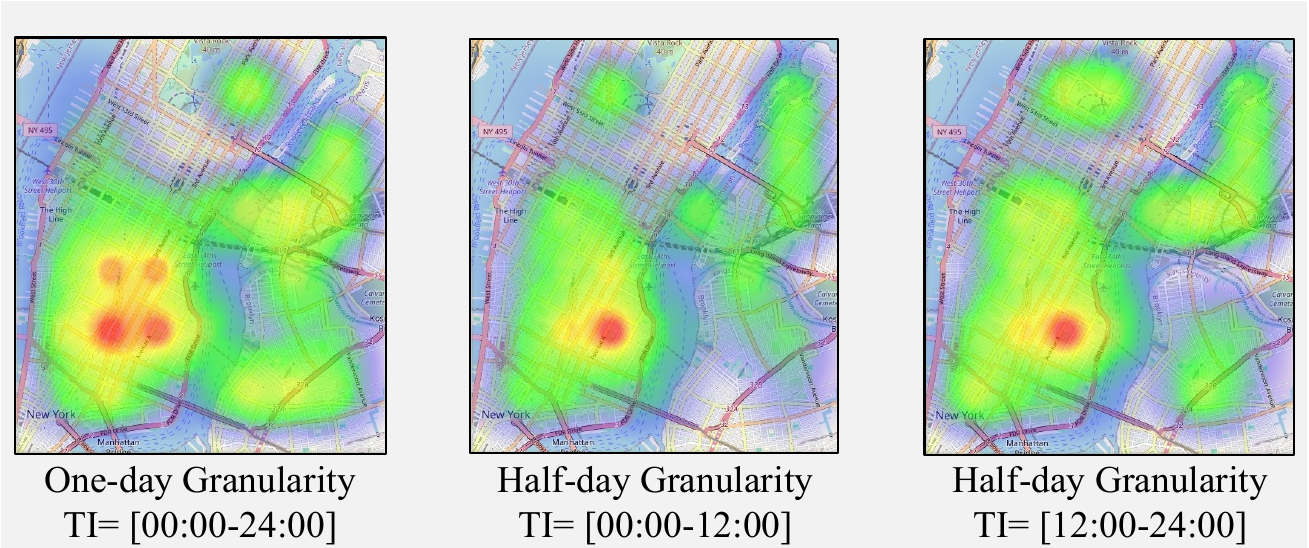}
      \label{fig:intro2}
    \end{minipage}
  }
  \caption{Comparison of true and predicted crime hotspots in New York City on Jan 31, 2018.  The crime hotspots across a city appear significantly different during various time intervals (TI). Existing methods usually predicted crime hotspots  with specified time granularities (day or half-day), which can result in inaccurate predictions if the predefined time intervals  used during training do not align with the  target time interval during prediction. }
  \label{fig:intro}
\end{figure}

\begin{example}
  Law enforcement agencies frequently analyze crime hotspots to develop predictive policing strategies~\cite{van2021right}. To optimize the allocation of police resources, these agencies often forecast crime hotspots for specific periods, such as ``what are the crime hotspots during the time interval [12:00-20:00]?'' According to Figure~\ref{fig:intro1}, in New York City, agencies can strategically focus their resources on the East Side of Manhattan during the period from 12:00 to 20:00. However, Figure~\ref{fig:intro1} also demonstrates that the distribution of crime hotspots varies significantly throughout the day. For instance, during the [06:00-18:00] interval, the hotspots are primarily located in Lower Manhattan. To predict these hotspots, existing methods typically train their models using predefined time granularities such as one-day intervals ([00:00-24:00]) or half-day intervals ([00:00-12:00] and [12:00-24:00]). However, these approaches can result in inaccurate predictions if the predefined time intervals—for example, a full day from [00:00-24:00] used during training—do not align with the specific target time interval, such as [12:00-20:00], during prediction.

  \label{example:ex1}
\end{example}

\noindent
{\bf Challenges.} This paper aims to predict crime hotspots with flexible time intervals, which is a significant but often overlooked problem. The primary challenge in this task arises from the sparsity and irregularities of multi-type crime events. Previous works have predominantly employed sequential models such as Long Short-Term Memory (LSTM), Gated Recurrent Unit (GRU), Transformer, etc., to address this problem \cite{hou2022integrated, ChaoHuang2018DeepCrimeAH, XianWu2020HierarchicallyST, cesario2024multi, 10.1145/3511808.3557350, 10.1145/3583780.3615027, DBLP:conf/dasfaa/HuLXTX24}. These models divide time into fixed intervals according to the predefined time granularities and aggregate events within each interval, but determining appropriate time granularities is challenging. Short intervals result in sparse data and poor performance, while long intervals may lead to a loss of necessary detail. Additionally, when the starting time and duration are flexible, the prediction targets are incompatible with the trained fixed-interval models, resulting in inaccurate predictions in real-world applications. Moreover, recent studies \cite{schoenberg2019recursive, okawa2019deep, mo2022predicting} have explored event-centric approaches, such as point processes \cite{schoenberg2019recursive}, for event prediction without the need for time interval division. However, incorporating correlations between different types of events and capturing spatiotemporal correlations presents a challenge for these approaches. Although neural point process methods have been developed to integrate comprehensive contextual information for event prediction, their complexity often hinders their effectiveness in modeling crime relationships.

\noindent
{\bf Presented work.} We introduce \MODEL, a novel event-centric framework for predicting crime hotspots across flexible time intervals. \MODEL~forecasts the probability of crime occurrence in urban areas during specified periods, identifying high-risk locations as potential crime hotspots. To achieve temporal flexibility in prediction, \MODEL~employs an innovative event-centric approach that incorporates two key components: a continuous-time attention network and a type-aware spatiotemporal point process, which captures crime-related features associated with individual time points for each target urban area. Based on the event-centric approach, \MODEL~then samples time points from the target interval and leverages the crime-related features of these sampled points to enable the flexible interval predictions.

\noindent
{\bf Contributions.} \MODEL~is unique in the following.
\begin{itemize}
  \item \ul{\it(Flexible interval prediction).}
        \MODEL~is a novel deep learning model for crime hotspots prediction with flexible time intervals, which does not require prediction granularity to be determined at training time.
        Unlike existing approaches that aggregate events within predefined time intervals, \MODEL~focuses on modeling event-centric crime features.
  \item \ul{\it(Continuous-time attention network).} \MODEL~uses a continuous-time attention network to learn crime context features for each city area at any time point. The crime context feature aggregated encodings from crime events that occurred in similar urban areas and timeframes. By considering relevant crime events, this module learned general crime knowledge about the target time and location. It also accounted for the multicategory and spatiotemporal correlations among crime events in a continuous-time manner, differing from existing methods.
            
  \item \ul{\it(Type-aware spatiotemporal point process). } \MODEL~incorporates a type-aware spatiotemporal point process designed to capture the evolving features of crime.  These features measure the risk of a specific type of crime at a given time and location. We design a type-aware continuous normalizing flow model to estimate the spatial conditional density function. We train this model using a likelihood-based approach to maximize the likelihood of specific type crime occurrences.
        
  \item \ul{\it(Performance evaluation).} The performance of \MODEL~is evaluated using real-world datasets through crime prediction tasks, ablation experiments, and hyperparameter experiments with multiple interval lengths and start times. The experimental results demonstrate that our model outperforms other methods by 16.53\% when applied to different flexible intervals for crime prediction.
\end{itemize}

The rest of the paper is structured as follows. We provide related work in Section \ref{sec:related work} and the preliminaries in Section \ref{sec:problem}. Then, we detail the \MODEL~in Section \ref{sec:methodology} and show experiment in Section \ref{sec:experiments}. Finally, the conclusion of the paper is presented in Section \ref{sec:conclusion}.

\section{Related Work}
\label{sec:related work}

\subsection{Deep Crime Prediction}
Crime prediction plays an instructive role in the field of security and protection, and has received widespread attention from social science and computer science \cite{zhao2018crime}. With the prominence of neural networks in various fields, there have been a number of crime prediction methods that use different neural networks to model spatio-temporal data \cite{jing2024deep}.  Techniques for solving crime prediction problem involve using convolutional neural networks \cite{wang2022hagen}, graph neural networks \cite{li2022spatial, LianghaoXia2021SpatialTemporalSH, 10.1145/3511808.3557657, 10.1145/3583780.3615065,10.1145/3583780.3614871}, attention networks \cite{Zhao_Liu_Cheng_Zhao_2022,XianWu2020HierarchicallyST,ChaoHuang2018DeepCrimeAH,huang2019mist,10.1145/3616855.3635764}, and large language models \cite{10508389, 10.1145/3583780.3615016}. 

For example, Wang et al. \cite{wang2022hagen} designed an end-to-end graph convolutional recurrent network to jointly capture the crime correlation between regions and the temporal crime dynamics by combining an adaptive region graph learning module with the diffusion convolution gated recurrent unit. Xia et al. \cite{LianghaoXia2021SpatialTemporalSH} employed a multi-channel routing mechanism with a hypergraph learning paradigm to model the impact of fine-grained semantic levels of cross-typed crime in a graph neural network framework. Huang et al. \cite{ChaoHuang2018DeepCrimeAH} proposed a hierarchical recurrent neural network with attention layers to capture dynamic crime patterns and explore interdependencies for crime prediction. Zhao et al. \cite{Zhao_Liu_Cheng_Zhao_2022} utilized a classification-labeled continuousization strategy and a weighted loss function for sparse classification problem, and propose a attention-based spatio-temporal multi-domain fusion network for crime prediction.Hu et al. \cite{10508389} proposed a hierarchical learning method that combines semantic and geographic information for sequential crime prediction.

However, although these works are effective, they all require datasets to be partitioned at fixed interval as input, and the prediction intervals need to be the same as the training intervals. This makes them unable to cope with the demand for flexible interval predictions. 

\subsection{Spatio-temporal Point Process}
A point process is an event-centric approach for discrete event modeling in probabilistic statistics. The spatio-temporal point process aims to predict continuous points in time and location and is commonly used in modeling earthquakes and aftershocks \cite{ogata1988statistical}, the occurrence and spread of wildfires \cite{Hering_Bell_Genton_2008}, epidemics and infectious diseases \cite{schoenberg2019recursive}. Recent works have explored neural network parameterizations of point process \cite{Du_Dai_Trivedi_Upadhyay_Gomez-Rodriguez_Song_2016, xiao2017modeling}, with an increasing focus on the dynamic relationships of spatiotemporal events\cite{0032DSJ023, pmlr-v162-yin22a,chen2020neural, zhou2022neural}.

For instance, Chen et al. \cite{chen2020neural} proposed a new parametric approach for spatio-temporal point processes, using neural ODEs as a computational method to model event data. Zhou et al. \cite{zhou2022neural} integrated a principled spatiotemporal point process with deep neural networks, which derives a tractable inference procedure by modeling the space-time intensity function as a composition of kernel functions and a latent stochastic process. Based on the powerful modelling capabilities of point processes, many research works have used point processes as a core component of predictive or generative task models \cite{yang2023contextual, long2023practical, DBLP:conf/aaai/JinLLH23, lin2024unified}. Yang et al. \cite{long2023practical} used a unique temporal point process to model non-fixed-length trajectories with continuous time distributions and accurately model the time in trajectory generation. 

However, crime events are very sparse and may be difficult to model directly and effectively.
Additionally, these works usually only take into account geolocation information, neglecting spatial information (geographical features, traffic, etc.).
Furthermore, by modelling towards a single type of event, these works fail to consider the interactions between different types of events, which are nonetheless very important in urban crime prediction.

\subsection{Continuous-time Sequence Models}
Continuous-time sequence models provide an elegant approach for describing irregular sampled time series, which can be considered as event-centric work by using irregularly sampled time series directly as input. Some works assume the latent dynamics are continuous and can be modeled by an ODE \cite{de2019gru, DupontDT19, kidger2020neural,NorcliffeBDML21}. And there are also some works designing specialised neural networks to model irregular time series data \cite{horn2020set, pmlr-v162-schirmer22a, ShuklaM21, chen2024contiformer,10.1145/3583780.3615239,10.1145/3583780.3614969, 10.1145/3589334.3645441}.

For example, De Brouwer et al. \cite{de2019gru} proposed GRU-ODE-Bayes to deal with non-uniformly sampled time series data, using a Poisson distribution process to explicitly model the probability of observation times. Kidger et al. \cite{kidger2020neural} designed the neural controlled differential equation model which operates directly on irregularly sampled and partially observed multivariate time series. Schirmer et al. \cite{pmlr-v162-schirmer22a} proposed continuous recurrent units to model temporal data with non-uniform time intervals in a principled manner, which incorporates a continuous-discrete Kalman filter into an encoder-decoder structure thereby introducing temporal continuity into the hidden state.
Shukla et al. \cite{ShuklaM21} designed an embedding of continuous time values and used attention mechanisms to adaptively weight observations at different time points to better capture dynamic changes in time series data. 

However, these works focus on prediction tasks over irregular time series without containing complex spatial information. This may make these techniques perform not well in spatiotemporal environments.
\begin{table}[t]
  \caption{Summary of Notations}
  \begin{center}
    \begin{tabular}{c|c}
      \hline
      \textbf{Symbol}                     & \textbf{Description}                                        \\
      \hline
      $m,n$                               & number of urban grid rows and columns                       \\
      $g$                                 & urban grid                                                  \\
      $t,s$                               & continuous time, continuous location                        \\
      $x,y$                               & longitude, latitude of location $s$                         \\
      $c$                                 & crime type                                                  \\
      $r$                                 & crime record triple $(t,s,c)$                               \\
      $d$                                 & dimension of a vector                                       \\
      $C$                                 & set of crime types                                          \\
      $R$                                 & set of crime records                                        \\
      $I$                                 & time interval                                               \\
      $\lambda$                           & instantaneous risk of crime                                 \\
      $\delta$                            & day of time $t$                                             \\
      $\widetilde{\cc}$                   & crime context feature of time interval $I$                  \\
      $\widetilde{V}$                     & crime evolving feature of time interval $I$                 \\
      $\mathbf{g}$                        & encoding of urban characteristics in each grid $g$          \\
      $\vt, \vs$                          & encoding of continuous time $t$ and continuous location $s$ \\
      $\vr$                               & encoding of target-aware event                              \\
      $\boldsymbol{r}$                    & encoding of record triple $r$                               \\
      $\boldsymbol{h}_t$                  & hidden state of $t$ in temporal intensity                   \\
      $\mathbf{W},\mathbf{Q}, \mathbf{K}$ & parameter matrices to be learnt of \MODEL                   \\
      
      \hline
    \end{tabular}
    \label{notation}
  \end{center}
\end{table}

\section{Preliminaries}
\label{sec:problem}
We define the crime hotspot prediction problem as follows.

\noindent
\textbf{City.}
As in previous works \cite{ChaoHuang2018DeepCrimeAH, LianghaoXia2021SpatialTemporalSH}, we represent the entire city space as an $m \times n$ matrix consisting of equal-sized grids.  Each grid represents an urban area. We associate each grid $g$  with a geolocation denoted as $s_g(x,y)$, representing the longitude and latitude of the grid center, respectively. To capture the urban features of each grid, such as points of interest and human flows, we employ region embedding models \cite{tang2022spemi} that utilize a vector $\mathbf{g} \in \mathbb{R}^d$ to encode the urban characteristics of each grid $g$. The dimensionality of the embedding vector is denoted as $d$, which we assume to be an even number in the following sections.

\noindent
\textbf{Crime record.} A crime record typically contains information about the date, time, and location of the crime event, as well as the type of crime committed. We define a crime record $r$ as a triple $(t, s, c)$, where $t$, $s$, and $c$ denote the datetime, geolocation, and type of the crime event, respectively. The value of $t$ indicates both the date and the time, allowing us to detect temporal connections at both the daily and hourly levels. 
% The grid $g$ is used to represent the location of the crime incident. 
Each crime records is mapped to a grid $g$ based on its location $s(x, y)$. We define the set of crime types as $C$, so that a crime type $c\in C$ conveys the semantic information of the crime event. We represent the overall set of crime records as $R$, and the set of crimes of type $c$ as $R_c$.

\noindent
\textbf{Crime hotspot prediction problem.} Given a set $R$ of crime records, a target time interval $I=[t^*, t^*+\Delta t]$, a target grid $g^*$, and a target crime type $c^*$, the crime hotspot prediction problem is to learn a model  to predict the probability $\mathbf{X}_{g^*,c^*}^{I} \in [0,1] $ that
the crime event with type $c^*$ is likely to occur in grid $g^*$ during the time interval $I$. 

\section{FlexiCrime: An Event-centric Framework}
\label{sec:methodology}

\subsection{Overview of Framework}
The architecture of \MODEL~is shown in Figure \ref{fig:methodology_framework}. To enable flexible-interval prediction, we focus on capturing crime-related features at specific time points rather than fixed intervals. In the flexible-interval prediction module, we sample $l$ time points $t_1, t_2, \ldots, t_l $ to collectively represent the features of the target time interval $I$. For each time point $t$, we design two types of features: first, crime context features $\widetilde{\cc}(t)$, which aggregate characteristics of historical crime events in similar areas and timeframes; second, crime evolving features $\widetilde{V}(t)$, which reflect the dynamic crime risk intensity at $t$. We concatenate these context and evolving features for each sampled time point, and to represent the crime features of the target interval $I$, we concatenate the features of all sampled time points as follows:

\begin{figure*}[t]
  \centering
    \includegraphics[width=\linewidth]{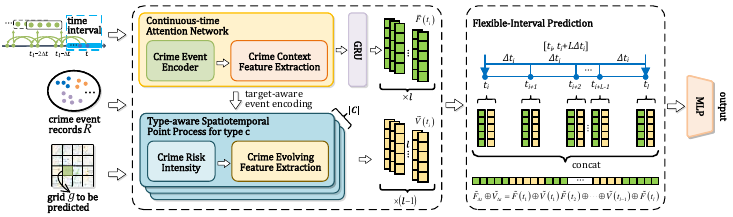}
    \label{fig:Architecture Overview}
  \caption{Overview of FlexiCrime}
  \label{fig:methodology_framework}
\end{figure*}

\begin{equation}
     \operatorname{concat}\left(\widetilde{\cc}({t_1}) , \widetilde{V}({t_1}),..., \widetilde{V}({t_{l-1}}), \widetilde{\cc}({t_l})\right)
\end{equation}

Based on this idea, \MODEL~utilizes two modules to capture crime context and evolving features: a continuous-time attention network and a type-aware spatiotemporal point process for each of the $|C|$ distinct crime types. These continuous-time crime features facilitate flexible interval prediction. The continuous-time attention network extracts crime context features at each time point, while the spatiotemporal point process models dynamic changes in crime risks between sampled points. These two types of modeling enhance the representation of temporal dynamics, ensuring a more comprehensive understanding of crime patterns. The following subsections provide detailed descriptions of these modules.

\subsection{Extracting Crime Context Feature with  Continuous-time Attention Network}
\begin{figure}[!t]
    \centering
    \includegraphics[scale=1.7]{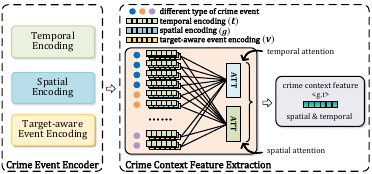}
    \caption{Continuous-time attention network}
    \label{fig:Continuous-time Attention Network}
\end{figure}

We introduce a continuous-time attention network (Figure~\ref{fig:Continuous-time Attention Network}) to aggregate historical crime records and extract crime context features. Spanning from time $t_i$ to time $t_j$, this network computes the crime context feature at any point within the interval $[t_i, t_j]$ by aggregating relevant events for a given target grid $g^*$ and target type $c^*$. This design allows \MODEL~to adaptively capture continuous-time crime information for each city grid. The network consists of two main components: a crime event encoder for event aggregation and a module for extracting crime context features.

\subsubsection{Target-aware Crime Event Encoder}
The crime event encoder processes individual crime events and generates corresponding embeddings, considering characteristics such as occurrence time, location, and crime specifics for a given target grid $g^*$ and target type $c^*$.

\noindent \textbf{Target-aware event encoding.}
We first encode the crime type and location features of the event. When predicting crime occurrences, it's essential to recognize that the relevance of historical crime events varies across different target locations \cite{behrouz2022cs}. For example, a historical crime in a prosperous area is more relevant to a well-developed target region than to an underdeveloped one. Additionally, the relevance of past events is influenced by the targeted crime type. Thus, for a given prediction task with target grid region embedding $\mathbf{g}^*$ and crime type $c^*$, we capture the target-relative encoding $\vr$ of each historical crime event $r(t,s,c)$ from both spatial and type relevance perspectives, as follows.

\begin{align}
    \label{eq:target-related-encoding}
    \vr & = \vr^{loc} || \vr^{type} \\
    \vr^{loc} & = \mathbf{W}^{loc}(\mathbf{g}^* || \mathbf{g}) \\
    \vr^{type} & = \mathbf{W}^{type}(\boldsymbol{c}^* || \boldsymbol{c}) \label{eq:typeencoding}
\end{align} 

The parameter matrices $\mathbf{W}^{loc}$ and $\mathbf{W}^{type}$ represent spatial and type relevances, respectively, while $||$ denotes vector concatenation. This concatenation enables the assessment of relevance between a crime event and both the target location and crime type. Each matrix has dimensions $\frac{d}{2} \times d^*$, where $d^*$ is the dimension of target-aware encoding. In Eq. \eqref{eq:target-related-encoding}, $\mathbf{g}$ is the region embedding for the grid containing the crime event, and $\boldsymbol{c}$~($\boldsymbol{c}^*$) encodes crime types $c$~($c^*$) using one-hot encoding.

With target-aware event encoding, we generate crime event representations that consider both the target location and crime type, distinguishing our model from existing approaches that overlook these factors. This targeted encoding captures more precise information for crime prediction.

\noindent \textbf{Temporal encoding.} 
We encode the time of day when a crime event occurs. Crime events can happen at any point in time. Since time is crucial for event description, it should not be discretized into fixed intervals (\eg~one-day periods). Instead, we adopt a continuous-time strategy \cite{ShuklaM21}, representing time $t$ as $\boldsymbol{t} \in \mathbb{R}^{d}$ for a more precise encoding:

\begin{equation}
    \vt[j] = 
    \begin{cases}
    \omega_{0} \cdot t+\alpha_{0}, & \text { if } j=0 \\ 
    \sin \left(\omega_{j} \cdot t+\alpha_{j}\right), & \text { if } 0<j<d
    \end{cases}
\end{equation}

\noindent \textbf{Spatial encoding.}
Similar to temporal encoding, we convert location information into spatial encoding. This spatial encoding captures the real geographical distance between crime events, allowing us to model their spatial near-by relationships effectively. The spatial encoding for location $s(x,y)$ can be represented as:
\begin{gather}
    \vs = \varphi(x) || \varphi(y)
\end{gather}
where $\varphi(\cdot)$ is a positional encoding function that outputs a vector of $\frac{d}{2}$ dimensions. We adopt the fixed positional encoding used in Transformer model \cite{AshishVaswani2017AttentionIA} and define the $j^{\text{th}}$ dimension of $\varphi(\cdot)$ as follows:

\begin{equation}
    \varphi(pos)[j]= 
    \begin{cases}
    \sin \left( pos / 10000^{2 k / (d/2)}\right), & \text { if } j=2k \\ 
    \cos \left( pos / 10000^{2 k / (d/2)}\right), & \text { if } j=2k+1
    \end{cases}
\end{equation}

Finally, the crime event $r$ is encoded as $\boldsymbol{r} =  (\vt, \vs, \vr)$. For simplicity, we assume that the dimensions for temporal, spatial, and target-aware encoding are all $d$, though users can adjust these dimensions as needed in our model.

\subsubsection{Crime Context Feature Extraction}
This module extracts crime context features $\cc_g(t)$ for each city grid $g$ at any time point $t$ within the historical crime record span $[t_i, t_j]$. It utilizes embeddings from the crime event encoder and considers temporal and spatial correlations among crime events. By aggregating relevant crime events and their embeddings, it generates a vector representation of the crime context for a specific time and location.

We compute crime context features by analyzing the relationship between the target grid location, target time, and historical crime events. We take into account the near-repeat phenomenon in sociology, which indicates that crimes are more likely to occur near previous events \cite{rahim2018vehicular}. To capture this, \MODEL~employs a proximity-based sampling approach, prioritizing crimes that occur close to the target grid and time. This method focuses on the most relevant events, reducing computational complexity during feature computation.

The proximity-based sampling method generates two sets of crime records, denoted as $R_0$ and $R_1$, which correspond to the grid $g$ centered at location $s$ and the time $t$, respectively. When considering a specific crime record $r=(t_r, s_r, c_r)$, its probability of being selected for $R_0$ and $R_1$ is denoted as $Pr(r | t)$ and $Pr(r | s)$, respectively. These probabilities are calculated using a softmax function, as shown below:
\begin{align}
Pr(r | t) &= \frac{EXP(-dis_0(t, t_r))}{\sum_{r' \in R} EXP(-dis_0(t, t_{r'}))} \\
Pr(r | s) &= \frac{EXP(-dis_1(s, s_r))}{\sum_{r' \in R} EXP(-dis_1(s, s_{r'}))}
\end{align}
Here, the functions $dis_0()$ and $dis_1()$ represent distance functions for time and location, respectively. While different distance metrics can be used based on requirements, we uniformly apply the Euclidean distance metric for simplicity. By using these softmax functions, records closer to the specified time point or location are more likely to be sampled.

\MODEL~computes the crime context feature $\cc_g(t)$ (denoted as $\cc(t)$) by aggregating target-aware crime embeddings of sampled records. The main idea is to assign weights to each crime record based on its relevance to grid $g$ and time $t$, emphasizing the most pertinent records. We utilize a multi-head attention mechanism($H$ heads)\cite{AshishVaswani2017AttentionIA} to calculate these weights. To illustrate the computation of the crime context feature, we take the $h^{th}$ attention head as an example:

\begin{equation}
\cc^h(t) = \sum_{r \in R_0} \mu (\vt, \vt_{r}) \vr_r + \sum_{r \in R_1} \eta (\vg, \vs_{r}) \vr_r
\label{eq:x_h1}
\end{equation}

In Eq. \eqref{eq:x_h1}, $\cc^h(t)$ represents the crime context feature for the $h^{th}$ attention head, obtained by aggregating target-aware crime embeddings $\vr_r$ from two sets of crime records: $R_0$ and $R_1$. The weight $\mu (\vt, \vt_{r})$ captures the temporal relationship between the temporal encoding $\vt$ of the target time $t$ and that of the crime event $r$ ($\vt_{r}$). Similarly, the weight $\eta (\vg, \vs_{r})$ reflects the spatial relationship between the spatial encoding $\vg$ of grid $g$ and that of the crime event $r$ ($\vs_{r}$), indicating the relevance based on spatial proximity. We utilize an attention mechanism to compute these weights.
\begin{equation}
\mu (\vt, \vt_{r}) = \frac{EXP ( \vt \mathbf{Q} (\vt_{r} \mathbf{K})^T /\sqrt{d / H} ) }{\sum_{r' \in R_0} EXP ( \vt \mathbf{Q} (\vt_{r'} \mathbf{K})^T /\sqrt{d / H} ) }
\end{equation}
Here, $\vt$ serves as a query vector, while $\vt_{r}$ acts as a key vector. We denote the query matrix as $\mathbf{Q}$ and the key matrix as $\mathbf{K}$. Similarly, the spatial importance weight $\eta$ is obtained using the corresponding query and key matrices:
\begin{equation}
\eta (\vg, \vg_{r}) = \frac{EXP ( \vg \mathbf{Q} (\vs_{r} \mathbf{K})^T /\sqrt{d / H}) }{\sum_{r' \in R_1} EXP ( \vg \mathbf{Q} (\vs_{r'} \mathbf{K})^T /\sqrt{d / H}) )}
\label{eq:alpha_S}
\end{equation}

By employing the attention mechanism, \MODEL~assigns appropriate weights to each crime record based on its temporal and spatial relationship to the target time and grid. This focus on relevant records enhances the representation of the crime context feature when aggregating the embeddings.

We represent the crime context $\cc_g(t)$ at any time point $t$ for grid $g$ as a concatenation of all attention heads, as shown in Eq. \eqref{eq:DCC}:
\begin{equation}
    \label{eq:DCC}
    \cc_g(t) = \operatorname{concat}([\cc^1(t), \cc^2(t), \cdots, \cc^H(t)]) \mathbf{W}^{MSA},
\end{equation}
where $\mathbf{W}^{MSA}$ is a learnable parameter matrix used to combine the outputs of multiple attention heads. As a result, we can compute the continuous-time crime contexts directly from discrete historical events, without the need for time interval divisions.

\subsection{Extracting Crime Evolving Feature with Type-aware Spatiotemporal
Point Process}

We extract crime-evolving features with a type-aware spatiotemporal point process. The type-aware spatiotemporal point process captures the dynamic nature of crime risk intensities and describes the evolution of crime-related features over time, which provides a continuous representation of spatiotemporal dynamics for a specific crime type.

Essentially, our crime evolving feature relies on the continuous-time intensity function in spatiotemporal point processes \cite{zhou2022neural}, which quantifies the instantaneous probability of an event occurring. 
Here, we consider $\lambda(t,s)$ as the instantaneous risk of crime for location $s$ at time $t$.
\begin{align}
    \lambda(t,s) =    \lim\limits_{\triangle t \to 0,\triangle s \to 0} \frac{Pr(t_i \in [t, t + \triangle t], s_i \in A(s,\triangle s))}{|\triangle t| |\triangle s|}
\end{align}
where $A(s,\triangle s)$ denotes the region of size $\triangle s$ centred on $s$. However, in the realm of crime prediction, the analysis of crime events also encompasses considerations of urban features and the correlation between different types of crimes.

\begin{figure}[t]
    \centering
    \includegraphics[scale=1.5]{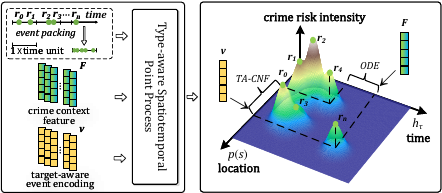}
    \caption{Type-aware spatiotemporal point process}
    \label{fig:Improved Spatiotemporal Point Process}
\end{figure}

\subsubsection{Crime Risk Intensity}
Our approach uses a type-aware intensity function to model the spatiotemporal risk intensity for crimes of a specific type $c^*$ occurring within a given day. To address the sparsity of type-$c^*$ crime events, we employ an event-packing strategy that modifies the datetime information $t=(\delta,\tau)$ of each crime record $r\in R$. 
Specifically, we remove the day portion $\delta$ and retain only the hours and minutes $\tau\in [00\!\!:\!\!00, 23\!\!:\!\!59]$, allowing us to focus on the time of day and disregard the specific date. We then sort the crime records by time $\tau$, packing them on a one-day scale, while also incorporating the features of the current date $\delta$ into the point process to capture the dynamic nature of crime risk throughout the day. This results in a crime risk intensity function denoted as $\lambda_{c^*}(\tau, s \mid R, \delta)$ in Eq. \eqref{eq:crime_risk}.

\begin{align}
\label{eq:crime_risk}
\lambda_{c^*}(\tau, s \mid R, \delta) = \lambda_{c^*}(\tau \mid R, \delta) p_{c^*}(s \mid \tau, R, \delta)
\end{align}

The function allows us to measure the level of crime risk at a particular location $s$ and time $(\delta, \tau)$, given the set of crime records $R$ and the specific day $\delta$. The function can be further broken down into two components. The first component, $\lambda_{c^*}(\tau \mid R, \delta)$, represents the conditional intensity of crime type $c^*$ occurring at time $\tau$, given the set of crime records $R$ and the specific day $\delta$, which captures the temporal dynamics and patterns of crime occurrence for the specific crime type within the given day. The second component, $p_{c^*}(s \mid \tau, R, \delta)$, represents the probability of crime type $c^*$ occurring at location $s$, given the time $\tau$, set of crime records $R$, and the specific day $\delta$. This component captures the spatial distribution and patterns of crime occurrence for the specific crime type at the given time and day. By multiplying these two components together, we obtain the overall crime risk intensity at location $s$ and time $(\delta, \tau)$. In the following, we will utilize $\lambda^*(\tau,s)$, $\lambda^*(\tau)$, and $p^*(s \mid\tau)$ to represent $\lambda_{c^*}(\tau,s \mid R, \delta)$, $\lambda_{c^*}(\tau \mid R, \delta)$, and $p_{c^*}(s \mid \tau, R, \delta)$ for the sake of brevity, respectively.

\subsubsection{Crime Evolving Feature Extraction}
As mentioned above, the crime context feature of discrete sampling points partially captures the characteristics of flexible time intervals. However, since the sampling points are discrete, it overlooks the time between adjacent points. To address this, we employ a type-aware spatiotemporal point process to model the time period features between these points, termed crime evolving features, which represent the information in the time period between adjacent sampling points.
To extract the crime evolving features, we estimate the temporal intensity $\lambda^*(t)$ and the spatial conditional density function $p^*(s\mid\tau)$ for a specific crime type $c^*$.

\noindent\textbf{Temporal intensity}. 
To obtain the temporal intensity $\lambda^*(t)$, we leverage the positive neural network $\psi_\lambda()$ that takes the hidden state $\boldsymbol{h}_t$ as input.  The hidden state $\boldsymbol{h}_t$ serves as a summary of the crime event history at any given time $t$, allowing us to make predictions about future events.

\begin{equation}
\lambda^*(\tau) = \psi_\lambda(\boldsymbol{h}_\tau)
\end{equation}

We utilize the capabilities of Ordinary Differential Equations (ODEs) to model and capture the continuous-time hidden state $\boldsymbol{h}_\tau$. This approach enables us to gain valuable insights into the dynamics of the hidden state as it evolves over time. Additionally, we incorporate instantaneous updates that capture sudden changes triggered by observed crime events. To compute the value of $\boldsymbol{h}_\tau$, we initiate the process with an initial state, denoted as $\boldsymbol{h}_0$, which is randomly sampled from a Gaussian distribution $\mathcal{N}$. This initial state serves as the starting point for the continuous evolution and instantaneous updates within the model. To express the continuous evolution and instantaneous updates, we employ the following equations:

\begin{align}
\boldsymbol{h}_0&\sim \mathcal{N}\\
\frac{d \boldsymbol{h}_\tau}{d \tau} & = f_t\left(\tau, \boldsymbol{h}_\tau,\delta \right) \\
\boldsymbol{h}_{\tau''} &= \boldsymbol{h}_{\tau'}+ \int_{\tau'}^{\tau''} f_t\left(\tau, \boldsymbol{h}_\tau,\delta\right) d \tau \label{eq:temporal-integral}\\
\lim_{\varepsilon \rightarrow 0} \boldsymbol{h}_{\tau_r+\varepsilon} & = \psi_t\left(\tau_r, \boldsymbol{h}_{\tau_r}, \boldsymbol{v}^{type}_r,\delta \right)
\end{align}

Here, $f_t()$ represents the rate of change of the hidden state over time, which takes into account the specific time of day $\tau$, the current hidden state $\boldsymbol{h}_\tau$, and day information $\delta$. The continuous evolution of the hidden state is determined by integrating the rate of change over a time interval, as shown in Eq. \eqref{eq:temporal-integral}. We incorporate instantaneous updates using $\lim_{\varepsilon \rightarrow 0} \boldsymbol{h}_{\tau_r+\varepsilon}$ to capture abrupt changes triggered by the observed crime event $r$ at time $\tau_r$. The parameter $\varepsilon$ indicates that $\boldsymbol{h}_\tau$ is a $c\grave{a}gl\grave{a}d$ function(left-continuous with right limits) and experiences discontinuous jumps modeled by $\psi_t()$. $\psi_t()$ describes immediate updates to the hidden state based on the event’s target-aware type encoding $\boldsymbol{v}^{type}_r$, which models cross-type correlations between event $r$ and the target crime type $c^*$ using Eq. \eqref{eq:typeencoding}. We parameterize $f_t()$  as a standard multi-layer fully connected neural network and use the GRU's update mechanism for $\psi_t()$.

\noindent \textbf{Spatial conditional density}.
We propose a type-aware continuous normalizing flow (TA-CNF) model to estimate the spatial conditional density function $p^*(s\mid\tau)$ for the specific crime type $c^*$. Based on jump CNF \cite{chen2020neural}, TA-CNF employs invertible transformations for flexible probabilistic modeling of spatio-temporal discrete events. To accurately estimate $p^*(s\mid\tau)$, we use ODEs to capture the continuous evolution of crime locations over time, which is similar to modeling temporal intensity. The equations for continuous evolution and instantaneous updates are as follows:

\begin{align}
s_0&\sim \mathcal{N}\\
\frac{d s_\tau}{d \tau} & =f_s\left(\tau, s_\tau, \boldsymbol{h}_\tau\right) \\
s_{\tau''} &= s_{\tau'}+ \int_{\tau'}^{\tau''} f_s\left(\tau, s_\tau, \boldsymbol{h}_\tau\right) d \tau \\
\lim s_{\tau_r+\varepsilon} & =\psi_s \left(\tau_r, s_{\tau_r}, \boldsymbol{h}_{\tau_r}, \cc_{g}(t_r)\right) \label{eq:spatial_update}
\end{align}

In Eq. \eqref{eq:spatial_update}, $\psi_s()$ takes the crime context feature $\cc_{g}(t_r)$ of grid $g$ as input, where $g$ represents the city grid for location $s_{\tau_r}$ and $t_r$ is the datetime of event $r$. Benefit from the target-aware crime event encoding, the crime context feature incorporates relationships between target type $c^*$, grid $g$, and time $t_r$, considering historical events. This continuous-time context enhances the CNF's ability to capture crime types and spatiotemporal correlations among crime events, differing from prior works that lack type-aware modeling.

Given the observed crime events at time $\bar{\tau}$, the TA-CNF  model  characterizes the probability distribution $p^*(s\mid\bar{\tau})$ of the spatial state $s$. In this model, the initial spatial state $s_0$ is sampled from a Gaussian distribution and serves as the starting point for estimating the spatial conditional density function $p(s\mid\bar{\tau})$. To capture the changes in density over time, the TA-CNF model establishes a continuous flow of transformations \cite{chen2018neural} through the use of Eq.~\eqref{eq:jump-CNF}, allowing for the integration of successive crime events.

\begin{equation}
\label{eq:jump-CNF}
\begin{aligned}
\log p^*\left({s} \mid \bar{\tau}\right) = \log {s}_0 +\underbrace{\int_{\tau_{k}}^{\bar{\tau}}-\text{Trace}\thinspace d \tau}_{\text {Change in density from last event to } t} +\underbrace{\sum_{r_i\in R, \tau_{i}<\bar{\tau}}\left(-\int_{\tau_{{i - 1}}}^{\tau_{i}} \text{Trace}\thinspace d \tau - \log \left|\operatorname{det} \frac{\partial \psi_s\left(\tau_{i}, {s}_{\tau_{i}}, \boldsymbol{h}_{t_{r_i}}\right)}{\partial s}\right|\right)}_{\text {Change in density up to last event}}
\end{aligned}    
\end{equation}
where $\operatorname{Trace}=\operatorname{tr}\left(\frac{\partial f_s\left(\tau, s_\tau, \boldsymbol{h}_\tau\right)}{\partial s}\right)$ represents the trace of the Jacobian determinant, $\tau_i$ denotes the time when the crime event $r_i$ occurs, and $r_k$ refers to the last event that happens before $\bar{\tau}$. 

\noindent\textbf{Likelihood maximization}. 
We use a likelihood-based approach to train the type-aware spatiotemporal point process, aiming to maximize the likelihood of type-$c^*$ crime occurrences. We consider the compound Poisson process \cite{daley2003introduction} and define the log-likelihood function as follows:
\begin{equation}
\label{eq:pp_loss}
\begin{aligned}
\log p(R_{c^*})=\sum_{r_i \in R_{c^*}} \log \lambda^*\left(\tau_i\right)-\int_0^T \lambda^*(\tau) d \tau +\sum_{r_i\in R_{c^*}} \log p^*(s_{\tau_i} \mid \tau_i)
\end{aligned}
\end{equation}
where $R_{c^*}$ is all the type-$c^*$ crime records. 
To train this module, we use gradient descent to maximum the log-likelihood. 
Additionally, we leverage neural ODE solvers to solve the ODEs involved in the training process \cite{NorcliffeBDML21}.

\noindent\textbf{Crime evolving feature}. The crime evolving feature can be represented by the instantaneous crime risk $\lambda^*(\tau,s)$ at a specific location and time. However, when considering the crime evolving features over a time interval and a city area, it becomes necessary to account for the integral of the instantaneous crime risk over both time and space, denoted as $\int\!\!\!\int\lambda^*(\tau,s)$. Since directly computing this integral can be computationally expensive, we employ a neural encoder to approximate and represent the crime evolving feature at a given time $\tau$ and day $d$ as follows:
\begin{equation}
\widetilde{V}({t}) = \operatorname{MLP}(\lambda^*(\tau, s), \boldsymbol{h}_\tau, \delta, \triangle t, \triangle s)
\end{equation}
Here, $\widetilde{V}(t)$ denotes the crime evolving feature at time $t$, where $t=(\tau, \delta)$. The parameters $\triangle t$ and $\triangle s$ represent the size of the time interval and the city grid, respectively. These parameters determine the granularity or resolution of the temporal and spatial dimensions within our model. In our approach, we employ a multilayer perceptron (MLP) as the encoder, which generates a $d$-dimensional vector as the output crime evolving feature.

\begin{figure}[!t]
    \centering
    \includegraphics[scale=0.85]{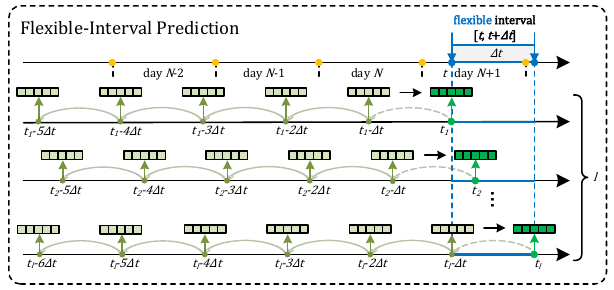}
    \caption{Flexible-interval prediction}
    \label{fig:sampling}
\end{figure}

\subsection{Flexible Interval Prediction}
\label{sec:flexible}
To perform flexible interval prediction, \MODEL~samples time points $t_1, t_2, \cdots, t_l$ within the time interval $I=[t_1,t_1+\Delta t]$, where $t_l=t_1+\Delta t$ is the end of $I$. The crime context and evolving features of these sampled time points are used to represent the time interval for crime hotspot prediction. 

However, the modeling of crime context features is limited to historical discrete events and cannot extend beyond the time of the last recorded crime event. To address this limitation, we use the GRU model to predict the  crime context features at the sampled time points in the future. As shown in Figure \ref{fig:sampling}, for each sampled time point $t_i$, we take a set of $L$ historical points, namely $t_i-\Delta t, t_i-2\Delta t, \cdots, t_i-L\Delta t$, into account. The GRU model utilizes the corresponding crime contexts at these historical points to generate the crime context at time $t_i$, as Eq. \eqref{eq:CCGRU}.
\begin{equation}
    \label{eq:CCGRU}
    \widetilde{\cc}(t_i) = GRU(\cc(t_i-\Delta t), \cc(t_i-2\Delta t),...,\cc(t_i-L\Delta t))
\end{equation}

By incorporating the crime evolving features, we represent the feature of the target time interval through the concatenation of the crime features, denoted as $\widetilde{\cc}_{\Delta t} \oplus \widetilde{V}_{\Delta t}$. This concatenation is achieved by combining the crime features at different time points within the interval. The resulting concatenated feature is used as input to an MLP model, followed by a sigmoid activation function, to predict the probability $\hat{\mathbf{X}}_{g^*,c^*}^{I}$ that a crime event of type $c^*$ is likely to occur in grid $g^*$ during the time interval $I$. Formally, this representation can be defined as follows:
\begin{align}
\label{eq:cc_feature1}
\widetilde{\cc}_{\Delta t} \oplus \widetilde{V}_{\Delta t} &=\widetilde{\cc}({t_1}) || \widetilde{V}({t_1}) || \widetilde{\cc}({t_2}) ||...|| \widetilde{V}({t_{l-1}})|| \widetilde{F}({t_l}) \\
\label{eq:out_1}
\hat{\mathbf{X}}_{g^*,c^*}^{I} & = \phi\left( \operatorname{MLP}(\widetilde{\cc}_{\Delta t} \oplus \widetilde{V}_{\Delta t})\right)
\end{align}
where the symbol $||$ represents the concatenation operation and $\phi$ is sigmoid activation function. 

\subsection{Training}
\label{sec:training}
This section introduces the loss function and  the learning process of \MODEL.

\subsubsection{Loss Function}
We  use the cross-entropy loss for the crime hotspot prediction, which can be formally represented as follows: 
\begin{equation}
\begin{aligned}
    \label{eq:l1}
    \mathcal{L}=&\dfrac{1}{|C|}\sum_{c^*\in C}
    \Big(- \dfrac{1}{|I^{set}|}\sum_{I \in I^{set}} \mathbf{X}^I_{c^*}\log \hat{\mathbf{X}}^I_{c^*}+\Big. \\
    &\Big.(1 - \mathbf{X}^I_{c^*}) \log (1-\hat{\mathbf{X}}^I_{c^*})\Big)+\xi\|\boldsymbol{\Theta}\|_2^2 
\end{aligned}
\end{equation}
where the set $I^{set}$ represents the training time intervals that need to be predicted and fitted. The model outputs the prediction result $\hat{\mathbf{X}}_{c^*}^I$, while the ground truth $\mathbf{X}_{c^*}^I$ is obtained from the actual dataset. In our approach, we evaluate the probability for every grid in the city when considering a specific crime type $c^*$. While the notations $\hat{\mathbf{X}}_{c^*}^I$ and $\mathbf{X}_{c^*}^I$ do not explicitly include the target grid $g^*$, this evaluation is performed for each grid within the city.

\begin{algorithm}[t]
    \renewcommand{\algorithmicrequire}{\textbf{Input:}}
    \renewcommand{\algorithmicensure}{\textbf{Output:}}
    \caption{Training Process of \MODEL~Framework}
    \begin{algorithmic}[1]
    \Require Crime event records $R$, city attributes $\{g\}$, learning rate $\gamma$, number of epochs 
    $n_1$,$n_2$,$n_3$.
    \Ensure Trained parameters 
 $\Theta_0$, $\Theta_1$, $\Theta_2$
    \LeftComment{$\Theta_0$: Parameters of continuous-time attention network }
    \LeftComment{$\Theta^c_1$: Parameters of  type-$c$ spatiotemporal point process}
    \LeftComment{$\Theta_2$: Parameters of flexible interval prediction module}
    \State Initialize all parameters $\Theta_0$, $\Theta^c_1$ for each crime type, $\Theta_2$

    \For{$epoch$ = 1 to $n_1$}
        \State Predict crime hotspot $\hat{\mathbf{X}}$ by using crime context features
        \State Calculate crime hotspots loss $\mathcal{L}$ according to Eq. \eqref{eq:l1}
        \State  Update parameters $\Theta_0$ and $\Theta_2$
    \EndFor
    \LeftComment{ Freeze parameters $\Theta_0$ and $\Theta_2$}
    \For{each crime type $c$}
        \For{$epoch$ = 1 to $n_2$}
            \State  Maximize the likelihood of type-$c$ crime occurrences
            \State Update parameters $\Theta^c_1$
        \EndFor
    \EndFor

    \LeftComment{ Freeze parameters $\Theta^c_1$ for each crime type}
    \For{$epoch$ = 1 to $n_3$}
        \State Predict crime hotspots $\hat{\mathbf{X}}$ by Eq. \eqref{eq:out_1}
        \State Calculate crime hotspot loss $\mathcal{L}$ according to Eq. \eqref{eq:l1}
        \State  Update parameters $\Theta_0$ and $\Theta_2$
    \EndFor
   
    \State \Return $\Theta_0$, $\Theta^c_1$ for each crime type, $\Theta_2$
    \end{algorithmic}
    \label{Alg:train}
\end{algorithm}

\subsubsection{Training Process} 

The spatiotemporal point process is trained using a three-step method based on the continuous-time attention network. In the first step, we train the attention network with the loss function from Eq. \eqref{eq:l1}, treating crime evolving features as zero, and then freeze its parameters. In the second step, we train the spatiotemporal point process for each crime type, optimizing it using the frozen parameters from the attention network. Finally, in the third step, we retrain the continuous-time attention network using the evolving features obtained from the second step, enabling \MODEL~to capture all relevant information for the target time interval.

The training process of \MODEL~is outlined in Algorithm \ref{Alg:train}. We initialize the parameters $\Theta_0$ of the attention network and $\Theta_2$ of the flexible interval prediction module (lines 2-5). Next, we maximize the likelihood in Eq. \eqref{eq:pp_loss} to train $\Theta_1$ of the spatiotemporal point process using the trained crime event encoder. Finally, we retrain $\Theta_2$ by incorporating crime context and evolving features from the predictions in Eq. \eqref{eq:out_1} (lines 6-10).

\MODEL's training is stage-wise, ensuring that each component is optimized step by step for optimal performance. Different optimization objectives necessitate distinct loss functions for the two components.
\section{Experiments}
\label{sec:experiments}
In this section, we evaluate the \MODEL~framework by conducting experiments on real-life urban crime datasets. Our aim is to answer the following research questions:

\begin{itemize}
  \item [\textbf{RQ1:}] How does \MODEL~compare to state-of-the-art baselines in predicting crime hotspots accurately?
        
  \item [\textbf{RQ2:}] What are the advantages of an event-centric approach for predicting crime hotspots using flexible time intervals?
        
  \item [\textbf{RQ3:}] How do different components of \MODEL~affect its performance?
  \item [\textbf{RQ4:}] Are the top-$k$ crime hotspots predicted by \MODEL~the true hotspots?
  \item [\textbf{RQ5:}] What does the predicted crime hotspots look like?
  \item [\textbf{RQ6:}] What are the impacts of hyperparameters on \MODEL's performance?
\end{itemize}

\subsection{Experiment Settings}
\begin{figure}[!t]
  \centering
  \subfigure[NYC Crime Ratio]{
    \begin{minipage}[t]{0.40\linewidth}
      \centering
      \includegraphics[width=\linewidth]{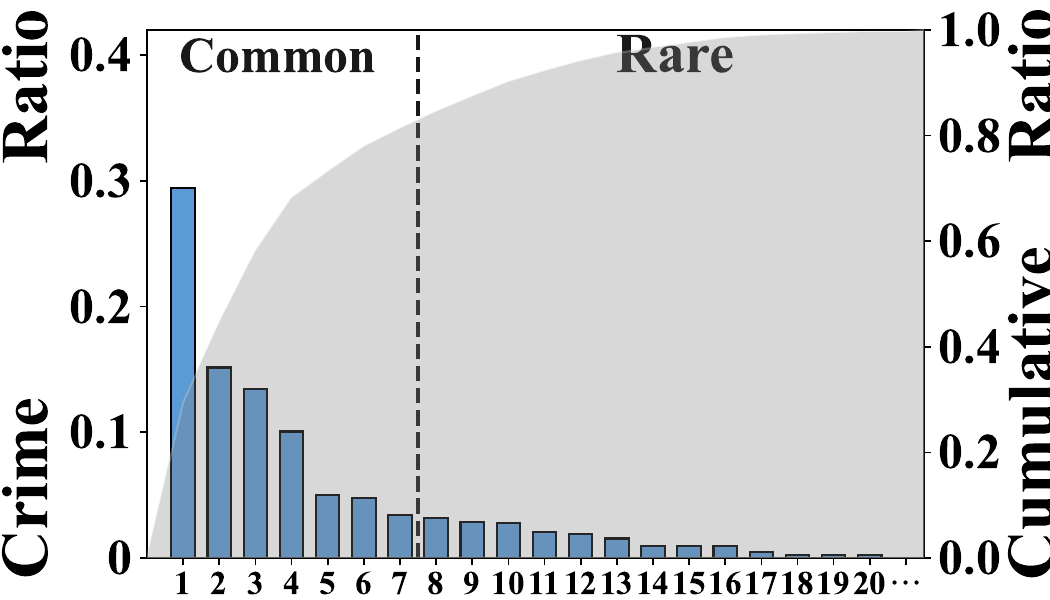}
      %\caption{fig1}
    \end{minipage}%
  }%
  \subfigure[SEA Crime Ratio]{
    \begin{minipage}[t]{0.40\linewidth}
      \centering
      \includegraphics[width=\linewidth]{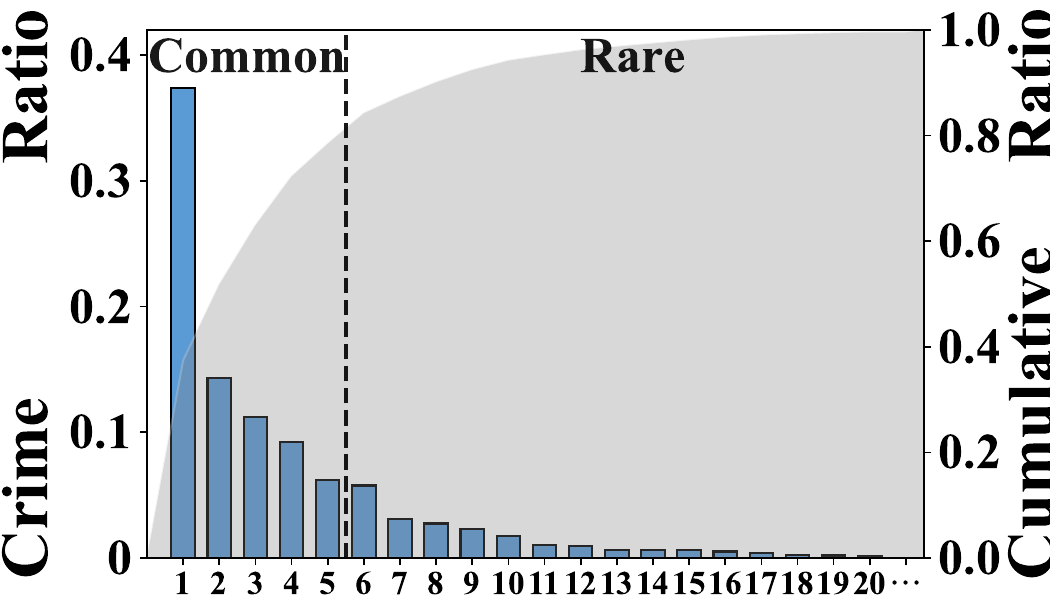}
      %\caption{fig2}
    \end{minipage}%
  }%
  \centering
  \caption{Crime ratio of various types (Indices 1, 2, 6, 9, 10, 11 represent theft, assault, robbery, weapons, sex crimes and forgery respectively in NYC; 1, 2, 3, 6, 7, 8 are theft, assault, burglary, fraud, trespass and robbery, respectively in SEA) }
  \label{tab:Crime Ratio}
\end{figure}

\noindent
\textbf{Datasets.}
We conducted experiments using two different types of urban datasets: Points of Interest (PoIs) and crime records, which were collected from New York City (NYC) and Seattle (SEA). The datasets used are as follows:

\begin{itemize}
  \item \textbf{PoI. } The PoIs in NYC and SEA were collected from Yelp~\cite{Yelp}, which are widely used PoI datasets. Each PoI is formatted as (venue name, category, address, latitude, longitude). The New York dataset contains 152 PoI categories, while the Seattle dataset contains 159 PoI categories.
  \item \textbf{Crime. }
        The crime records in NYC and SEA were collected from NYC OpenData\cite{nypd_complaint_data} and Seattle OpenData\cite{spd_crime_data}, respectively.
        Each record is formatted as (crime category, timestamp, and geographical coordinates). We use a cumulative ratio of 80\% as a cutoff to divide the crimes into common and rare categories, as shown in Figure \ref{tab:Crime Ratio}. We then select three common and three rare crimes from each dataset separately (see Table \ref{tab:dataset detail}).
\end{itemize}

\noindent
The grid-based mapping strategy partitions NYC and SEA into disjoint geographical grids of 1 km × 1 km. For crime records, \MODEL~utilizes all types of crime data as input for prediction, not just the target types. We applied various divisions and time interval sizes to create the training and test datasets, maintaining a 7:1 ratio.

\begin{table}[!t]
  \centering
  \caption{Details of the crime datasets}
  
  \begin{tabular}{c|c|c||c|c}
    \hline
    Data  & \multicolumn{2}{c||}{NYC-Crimes} & \multicolumn{2}{c}{SEA-Crimes} \\
    \hline
    Time Span & \multicolumn{2}{c||}{Jan, 2016 to Dec, 2017} & \multicolumn{2}{c}{Jan, 2020 to Dec, 2021} \\
    \hline
               & Theft      & Assault  & Theft    & Assault \\
               & (277555)   & (144883) & (50448)  & (21291) \\
    \cline{2-5}
    Category   & Robbery    & Weapons  & Burglary & Fraud   \\
    (Cases \#) & (29507)    & (19141)  & (20259)  & (14180) \\
    \cline{2-5}
               & Sex Crimes & Forgery  & Trespass & Robbery \\
               & (15844)    & (11558)  & (4447)   & (3238)  \\
    \hline
  \end{tabular}%
  \label{tab:dataset detail}%
\end{table}%

\noindent
\textbf{Evaluation metrics.}~We evaluate the performance of all compared methods in predicting crime hotspots in each region of a city using two types of evaluation metrics. The F1-score, which balances precision and recall, is employed to assess the accuracy of predicting specific categories of crime hotspots. Additionally, we utilize Marco-F1 and Micro-F1~\cite{tang2009large} to evaluate the prediction accuracy across different crime categories. These widely-used metrics are well-suited for multi-class problems.

\noindent\textbf{Baseline methods.}
We evaluate the performance of \MODEL~by comparing it with nine different prediction techniques, including three time series prediction methods (SVM, ARIMA, mTAND), one point process method (NSTPPs), and five deep crime prediction methods (DeepCrime, STtrans, ST-SHN, ST-HSL, AttenCrime). Here is a brief description of each method:

\begin{itemize}
  \item \textbf{SVM}~\cite{Chang_Lin_2012} is a time series prediction method that uses a kernel function to non-linearly map the input data to a high-dimensional feature space.
        
  \item \textbf{ARIMA}~\cite{pan2012utilizing} is a time series analysis method that measures the strength of the relationship between a dependent variable and other varying variables in the time series data.
        
  \item \textbf{mTAND}~\cite{ShuklaM21} learns embeddings of continuous time values and utilizes attention mechanisms to generate time series with a variable number of observations.
        
  \item \textbf{NSTPPs}~\cite{chen2020neural} is a point process method that parameterizes spatio-temporal point processes using Neural ODEs, allowing for high-fidelity models of discrete events localized in continuous time and space.
        
  \item \textbf{DeepCrime}~\cite{ChaoHuang2018DeepCrimeAH} uses recurrent neural networks to encode the periodicity of crime, and attention mechanisms are employed to aggregate criminal features.
        
  \item \textbf{STtrans}~\cite{XianWu2020HierarchicallyST} explores sparse crime data through a two-stage network of spatio-temporal transformers, encoding relationships between unbalanced multidimensional spatio-temporal data.
        
  \item \textbf{ST-SHN}~\cite{LianghaoXia2021SpatialTemporalSH} performs crime message passing by designing a temporal and spatial messaging architecture with an integrated hypergraph learning paradigm.
        
  \item \textbf{ST-HSL}~\cite{li2022spatial} encodes regional crime dependencies in the entire urban space using cross-regional hypergraph structure learning, and complements sparse crime representations with self-supervised learning.
        
  \item \textbf{AttenCrime}~\cite{Zhao_Liu_Cheng_Zhao_2022} employs a classification-labeled continuousization strategy and a weighted loss function for sparse classification problems. It proposes an attention-based spatio-temporal multi-domain fusion network for crime prediction.
        
\end{itemize}

\noindent\textbf{Hyper-parameter settings.} In our experiments, we set the size of spatial encoding, temporal encoding, and target-related encoding to 64. The hidden size is also set to 64, and the number of heads in the multi-head attention mechanism is set to 4. We sample 4 points and use 32 reference points. The batch size is set to 48, while the learning rate and weight decay are set to 0.0005 and 0.00005, respectively. By default, the time interval (TI), time granularity (TG), and starting time are set to [00:00–24:00], 24 hours, and 00:00, respectively.

\begin{table*}[tbp]
  \small
  \centering
  \renewcommand\arraystretch{1.05}
  \caption{Crime hotspots prediction performance  (TI=[00:00-24:00]) on NYC and SEA datasets in terms of Micro-F1 and Macro-F1~(RQ1)}
  \resizebox{\linewidth}{!}{
    \begin{tabular}{c|*{2}{c}|*{2}{c}|*{2}{c}|*{2}{c}|*{2}{c}|*{2}{c}}
      \toprule
      \multirow{3}[6]{*}[1.7ex]{Model} & \multicolumn{6}{c|}{NYC} & \multicolumn{6}{c}{SEA} \\
      \cline{2-13} & \multicolumn{2}{c|}{Common Crimes} & \multicolumn{2}{c|}{Rare Crimes} & \multicolumn{2}{c|}{Overall} & \multicolumn{2}{c|}{Common Crimes} & \multicolumn{2}{c|}{Rare Crimes} & \multicolumn{2}{c}{Overall} \\
      \cline{2-13} & Micro-F1         & Macro-F1         & Micro-F1         & Macro-F1         & Micro-F1         & Macro-F1         & Micro-F1         & Macro-F1         & Micro-F1         & Macro-F1         & Micro-F1         & Macro-F1         \\
      \cline{1-13}
      SVM          & 0.4563           & 0.3864           & 0.2090           & 0.1893           & 0.4147           & 0.2879           & 0.3438           & 0.3363           & 0.0682           & 0.0724           & 0.2335           & 0.2044           \\
      
      ARIMA        & 0.4716           & 0.4027           & 0.1010           & 0.0962           & 0.2494           & 0.2406           & 0.3968           & 0.3887           & 0.0712           & 0.0388           & 0.2432           & 0.2138           \\
      
      mTAND        & 0.3842           & 0.4154           & 0.1344           & 0.0985           & 0.2809           & 0.2569           & 0.3473           & 0.3282           & 0.0735           & 0.0701           & 0.2387           & 0.1992           \\
      
      NSTPPs       & 0.3818           & 0.3731           & 0.1853           & 0.2144           & 0.2741           & 0.2937           & 0.3401           & 0.3318           & 0.0723           & 0.0826           & 0.2183           & 0.2072           \\
      
      DeepCrime    & 0.4918           & 0.4181           & 0.2183           & 0.1996           & 0.4086           & 0.3089           & 0.3950           & 0.3665           & 0.1535           & 0.1281           & 0.3389           & 0.2473           \\
      
      STtrans      & 0.5110           & 0.4517           & 0.2240           & 0.2189           & 0.4366           & 0.3353           & 0.4046           & 0.3938           & 0.1875           & \ul{0.1747}      & 0.3370           & 0.2842           \\
      
      ST-SHN       & 0.5508           & 0.4886           & 0.2371           & 0.2266           & 0.4166           & 0.3576           & \ul{0.4194}      & 0.3834           & 0.1702           & 0.1399           & 0.3658           & 0.2617           \\
      
      ST-HSL       & 0.4981           & \ul{0.5059}      & 0.2246           & \ul{0.2543}      & 0.4176           & \ul{0.3801}      & 0.3927           & 0.3841           & 0.1838           & 0.1662           & 0.3000           & 0.2751           \\
      
      AttenCrime   & \ul{0.5812}      & 0.4982           & \ul{0.2631}      & 0.2490           & \ul{0.4412}      & 0.3736           & 0.4104           & \ul{0.4236}      & \ul{0.1901}      & 0.1620           & \ul{0.3851}      & \ul{0.2928}      \\
      
      \midrule
      FlexiCrime   & \textbf{0.6447 } & \textbf{0.5488 } & \textbf{0.3196 } & \textbf{0.2811 } & \textbf{0.5412 } & \textbf{0.4149 } & \textbf{0.4890 } & \textbf{0.4765 } & \textbf{0.1945 } & \textbf{0.1866 } & \textbf{0.4231 } & \textbf{0.3315 } \\
      \bottomrule
    \end{tabular}%
  }
  \label{tab:prediction performance}%
\end{table*}%

\subsection{Overall Perfomance (RQ1)}
\label{Sec:Perfomance_comparison}
To evaluate the performance of all the compared methods in predicting crime hotspots for the next day, we used the Marco-F1 and Micro-F1 metrics. The prediction time interval considered was 24 hours, starting at 00:00. In contrast to the prior studies, \MODEL~takes discrete events as input, eliminating the need for partitioning crime data based on time intervals. Table \ref{tab:prediction performance}
present the performance comparison between \MODEL~and different baseline methods for crime prediction.

These substantial enhancements can be attributed to two key factors. First, our model focuses on crime events and does not require manual setting of the time interval of input data. It dynamically captures the impact and spatial dependencies among different types of crime events, fully exploring their correlations. This approach enriches the prediction information, resulting in more abundant and accurate predictions. Second, our  design incorporates continuous-time crime contexts and evolving features to better represent interval crime features, which enables more accurate performance, even with the sparse and rare crimes.

\subsection{Results when Changing Time Intervals (RQ2)}
\label{Sec:rq2}
\noindent \textbf{Changing start time and length of prediction interval.}
\MODEL~offers the advantage of flexible time interval selection.  To validate this point, we trained various models on specific datasets characterized by a consistent 24 hours time interval, starting from 0:00. Then, we systematically conducted comparative experiments by adjusting the prediction start time and interval sizes to evaluate their impact on prediction accuracy.

\begin{figure}[t]
  \centering
  \begin{minipage}[t]{\linewidth}
    \centering
    \includegraphics[width=0.6\linewidth]{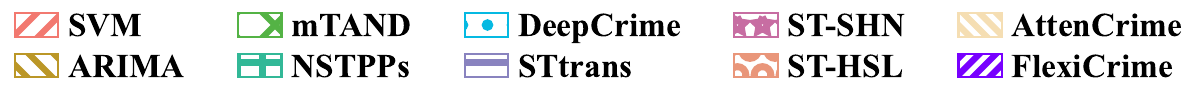}
  \end{minipage}%
  \\
  \centering
  \subfigure[NYC Micro-F1 Decrease]{
    \begin{minipage}[t]{0.24\linewidth}
      \centering
      \includegraphics[width=\linewidth]{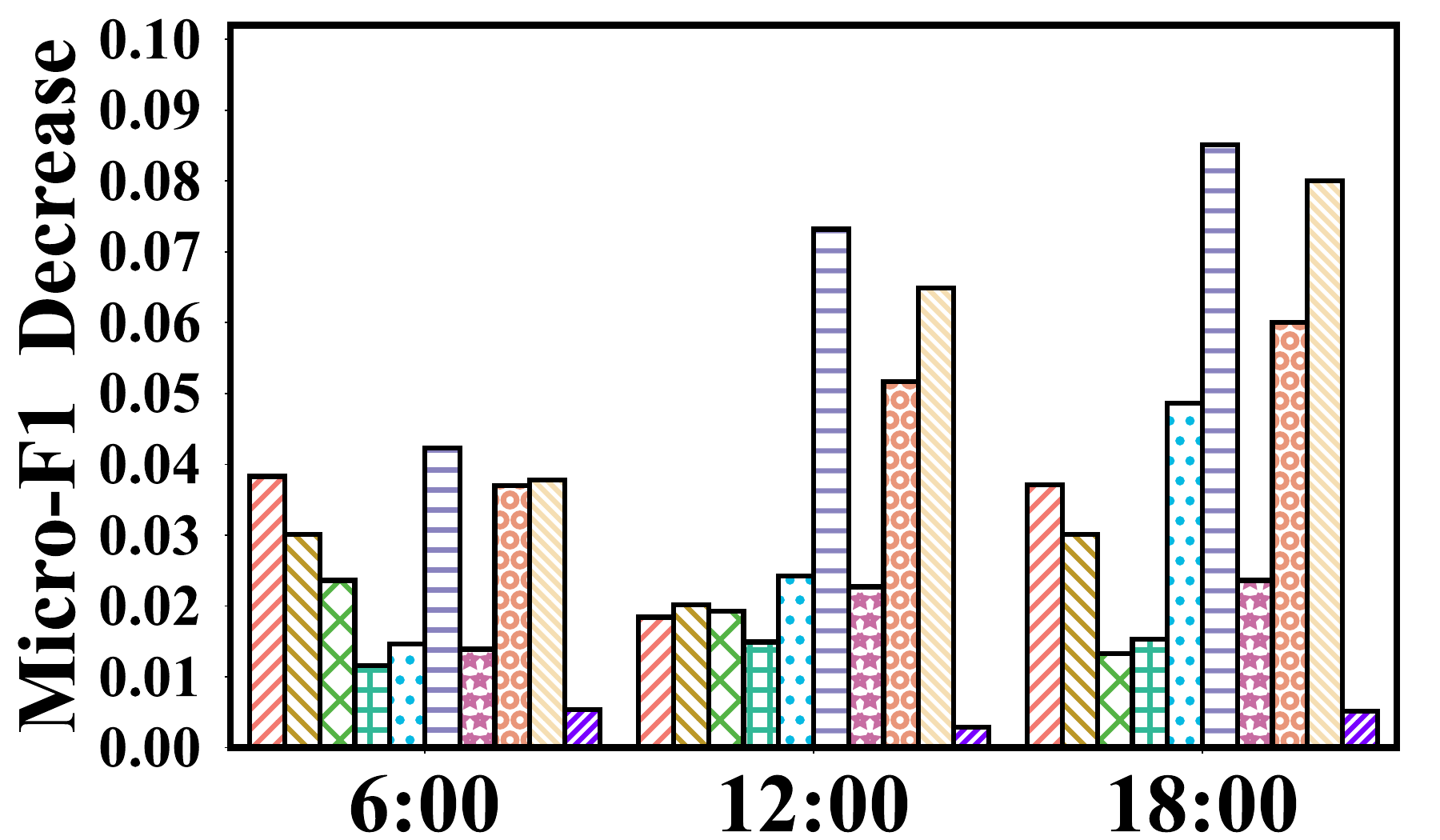}
    \end{minipage}%
  }%
  \subfigure[NYC Macro-F1 Decrease]{
    \begin{minipage}[t]{0.24\linewidth}
      \centering
      \includegraphics[width=\linewidth]{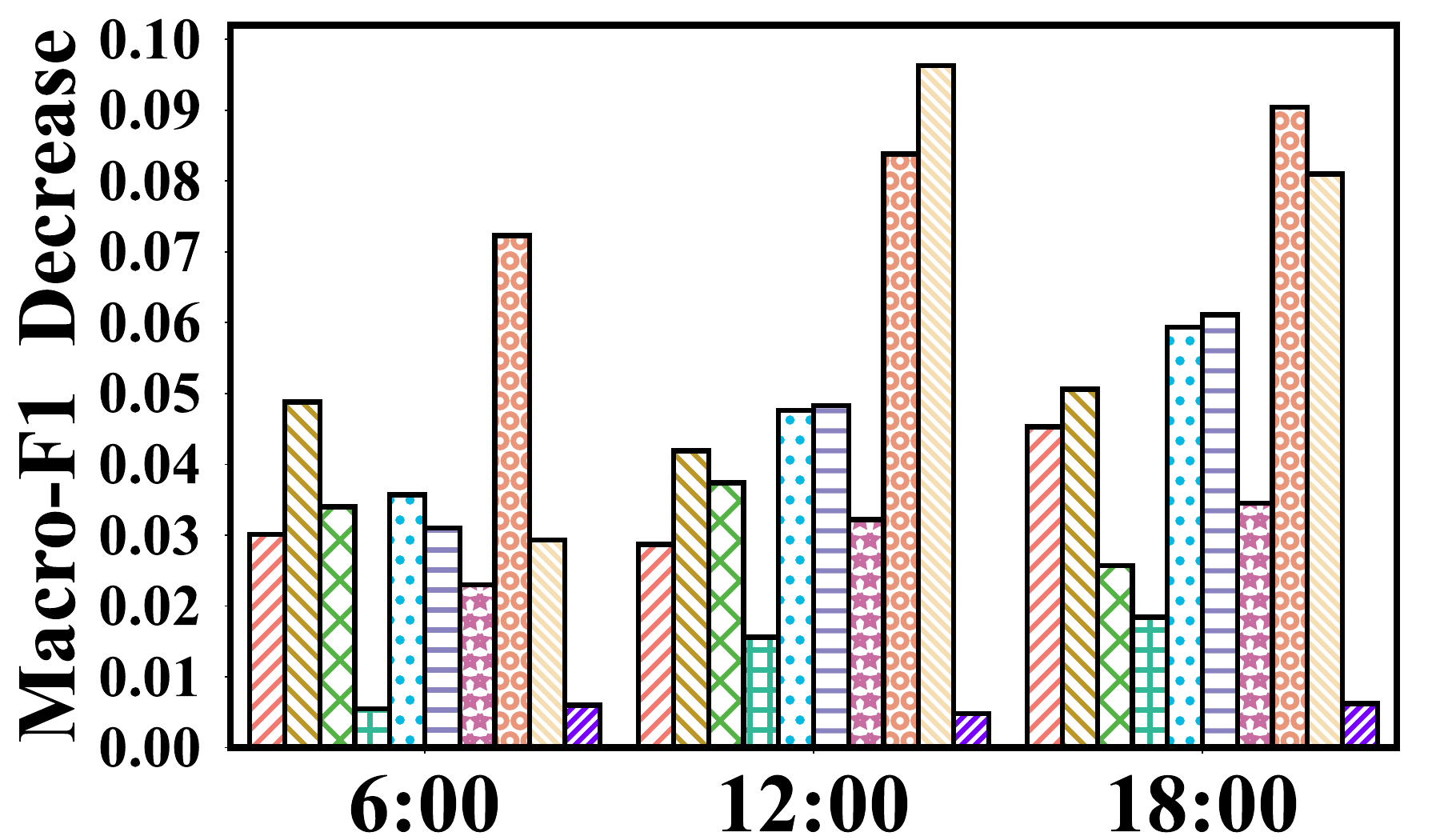}
    \end{minipage}%
  }
  \subfigure[SEA Micro-F1 Decrease]{
    \begin{minipage}[t]{0.24\linewidth}
      \centering
      \includegraphics[width=\linewidth]{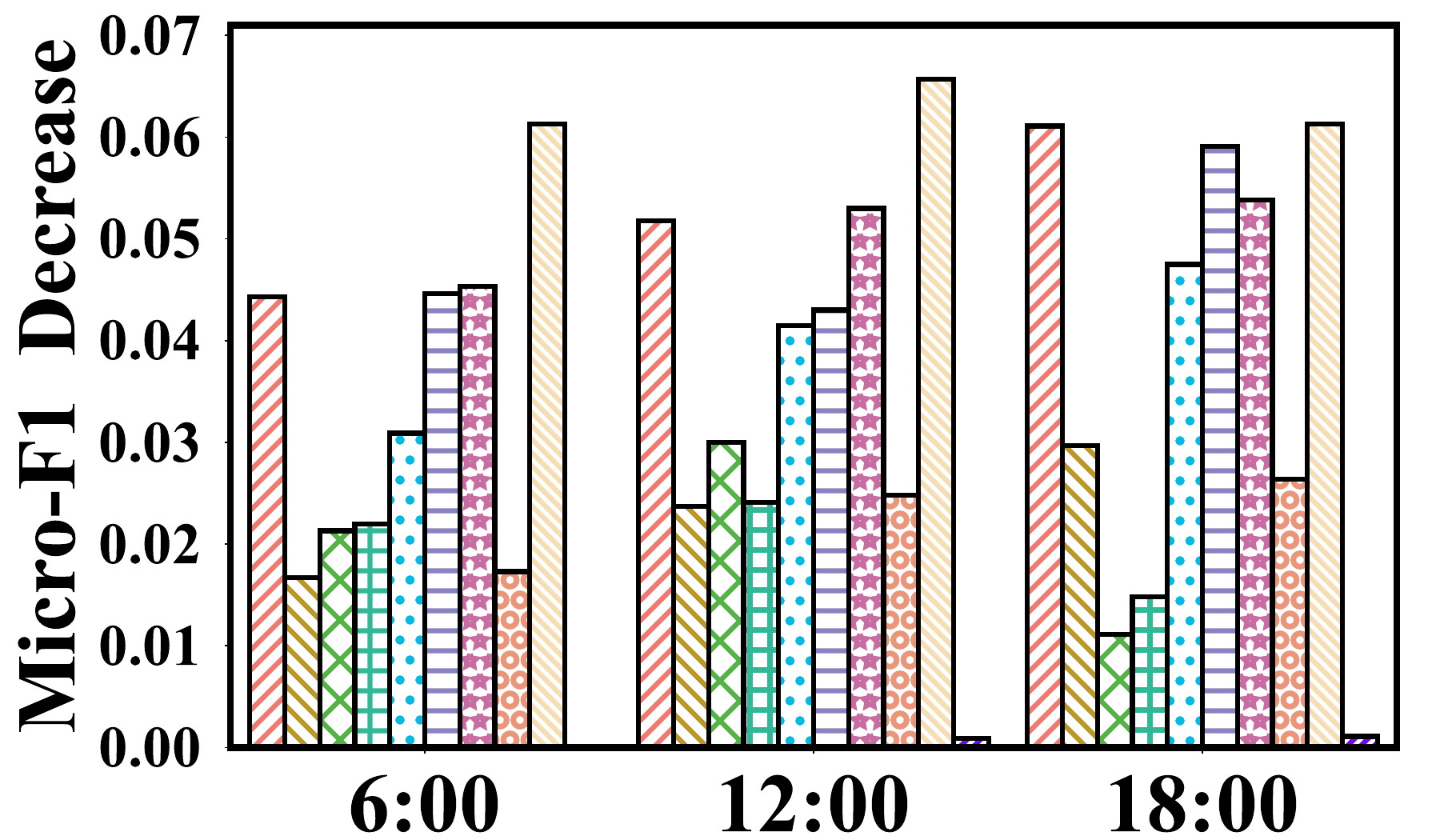}
    \end{minipage}%
  }%
  \subfigure[SEA Macro-F1 Decrease]{
    \begin{minipage}[t]{0.24\linewidth}
      \centering
      \includegraphics[width=\linewidth]{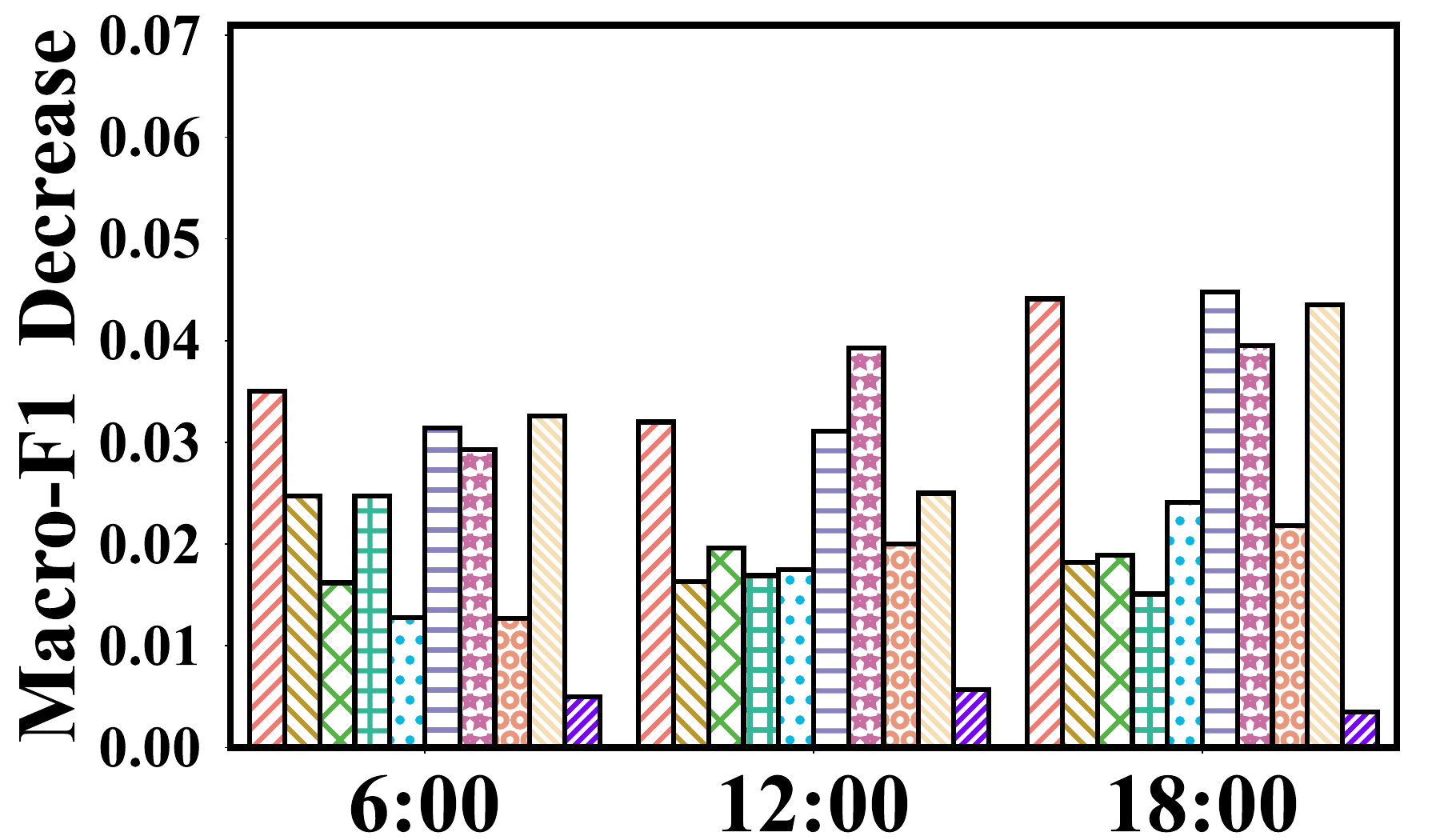}
    \end{minipage}%a
  }%
  \caption{Micro-F1 and Macro-F1's decrease with different starting times (TG=24 hours) among different methods~(RQ2)}
  \label{tab:decrease Result}
\end{figure}

The decreases for each scenario are visualized in Figure \ref{tab:decrease Result}. We calculated them by comparing the performance to the overall results in Table \ref{tab:prediction performance} (start time: 00:00, time interval size: 24 hours). Figure \ref{tab:decrease Result} shows that all baseline methods exhibit a significant performance decrease when the prediction start time varies. For example, ST-SHN shows an average decrease of 13.8\% in SEA prediction, and ST-HSL shows a 21.6\% decrease in NYC predictions. In contrast, \MODEL~experiences only a slight fluctuation of less than 0.05\%.

\begin{figure}[t]
  \centering
  \begin{minipage}[t]{\linewidth}
    \centering
    \includegraphics[width=.8\linewidth]{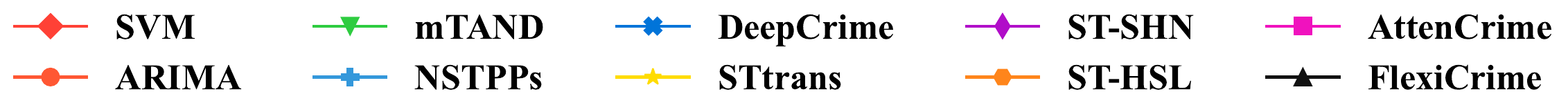}
  \end{minipage}
  \\
  \subfigure[NYC Macro-F1 Result]{
    \begin{minipage}[t]{0.24\linewidth}
      \centering
      \includegraphics[width=\linewidth]{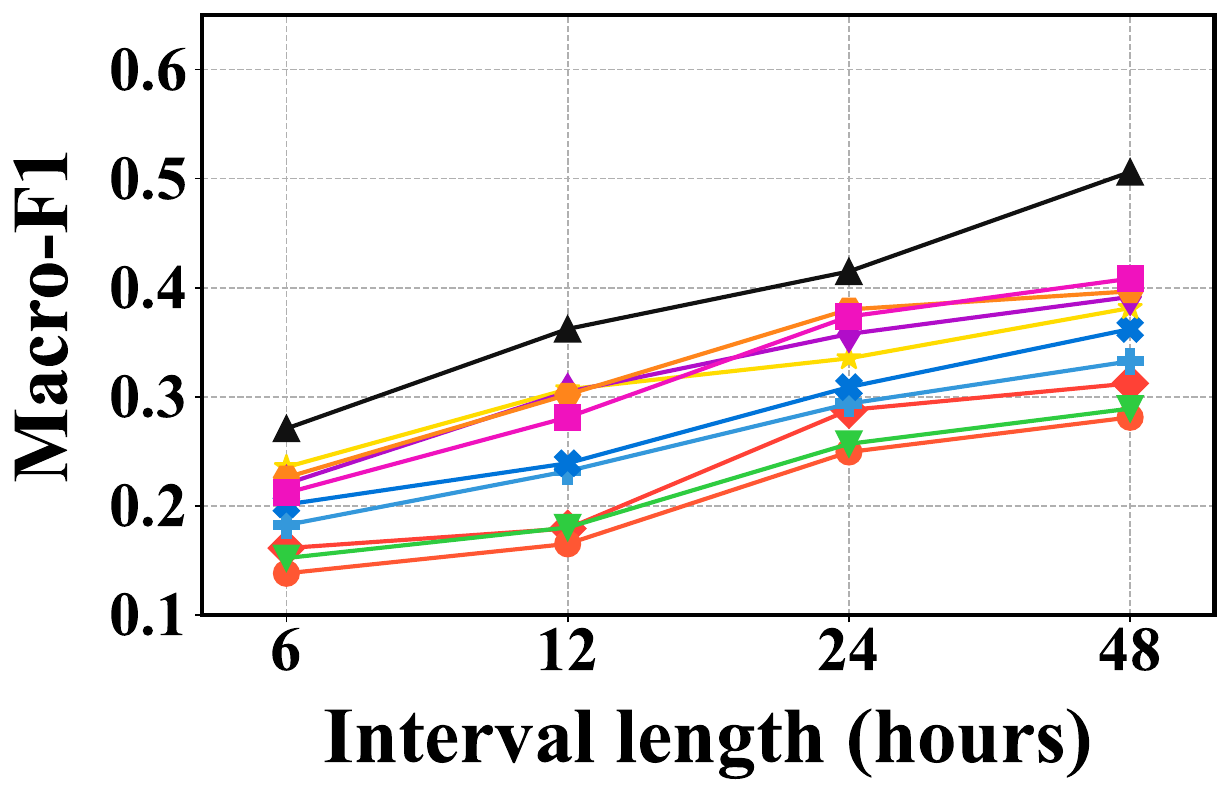}
    \end{minipage}%
    \setcounter{subfigure}{1}\renewcommand{\thesubfigure}{(\alph{subfigure})}
  }%
  \subfigure[NYC Micro-F1 Result]{
    \begin{minipage}[t]{0.24\linewidth}
      \centering
      \includegraphics[width=\linewidth]{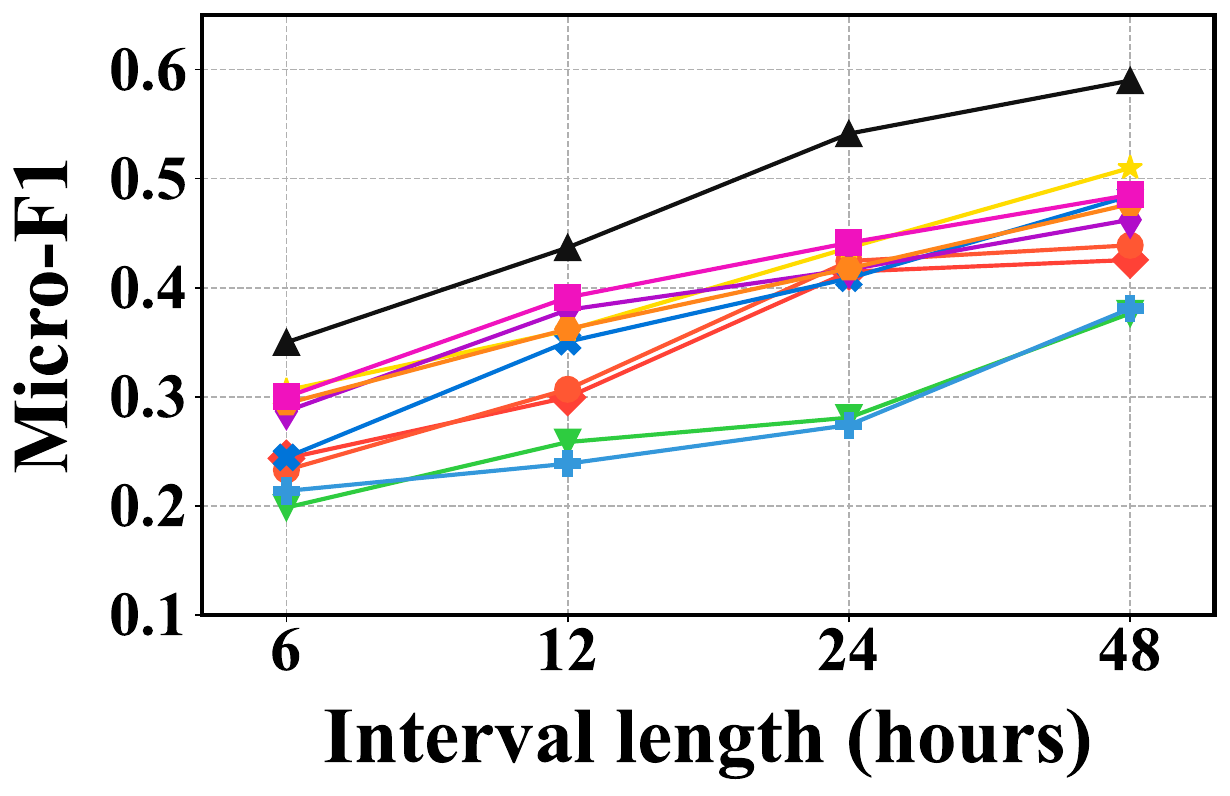}
    \end{minipage}%
  }
  \subfigure[SEA Macro-F1 Result]{
    \begin{minipage}[t]{0.24\linewidth}
      \centering
      \includegraphics[width=\linewidth]{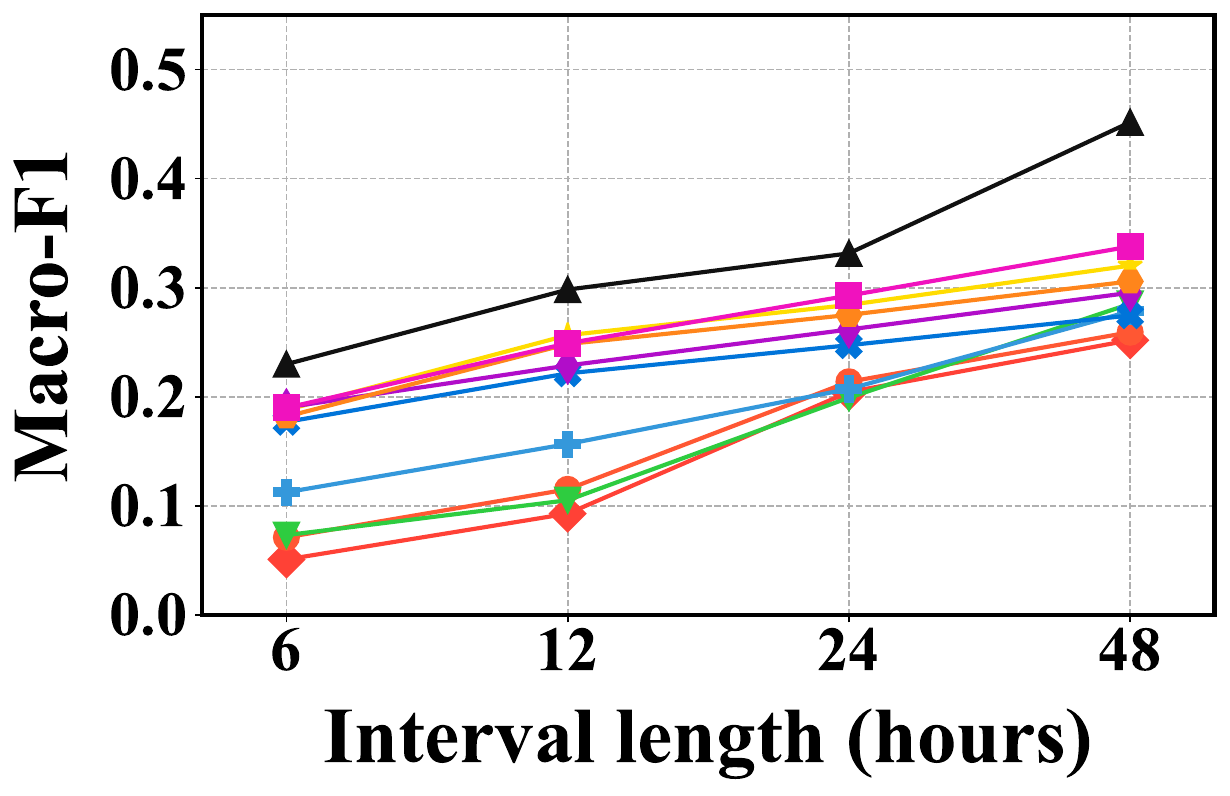}
    \end{minipage}%
  }%
  \subfigure[SEA Micro-F1 Result]{
    \begin{minipage}[t]{0.24\linewidth}
      \centering
      \includegraphics[width=\linewidth]{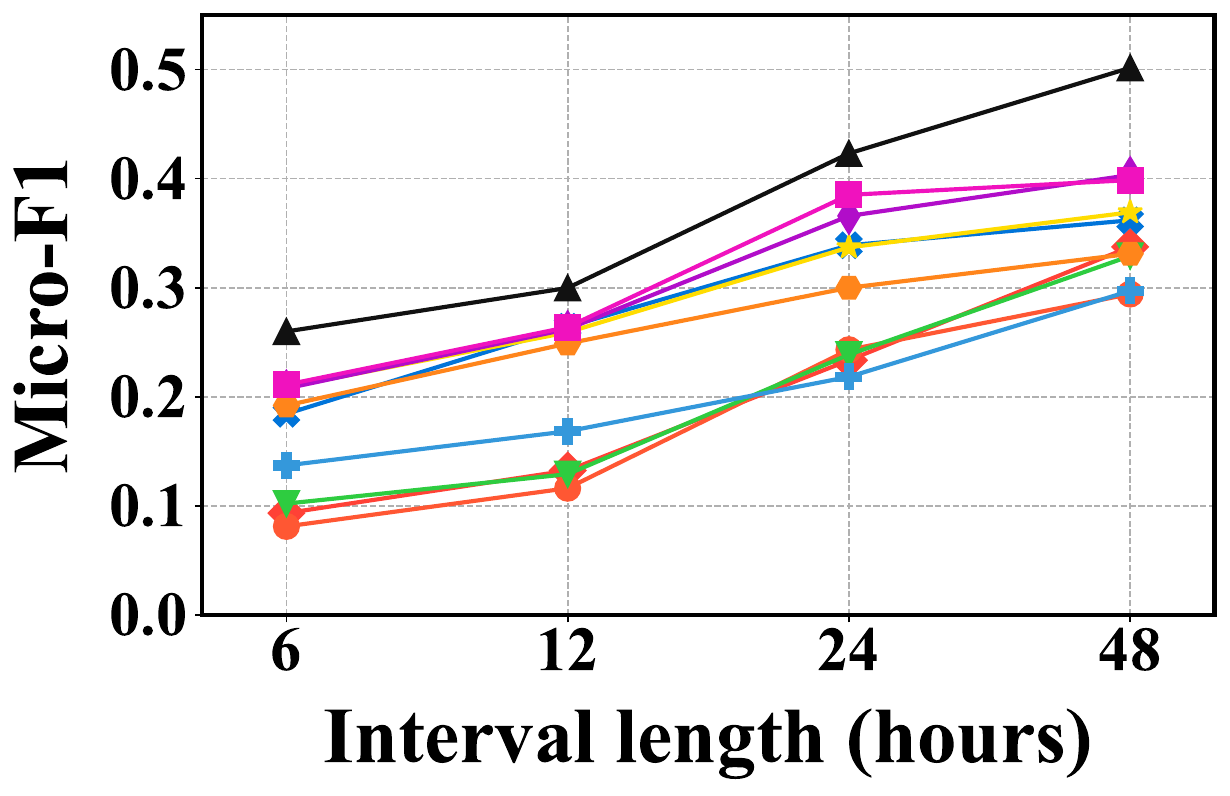}
    \end{minipage}%
  }%
  \centering
  \caption{Results when changing length of time interval~(RQ2). The baseline models were trained on 24-hour granularities across four different interval lengths.}
  \label{tab:multi_granularity Result}
\end{figure}

We also tested \MODEL's performance at different prediction time intervals, as shown in Figure~\ref{tab:multi_granularity Result}. Compared to baseline models trained on 24-hour granularities at four prediction interval sizes, \MODEL~consistently outperformed others, with the most significant improvement at the 48-hour interval. On average, \MODEL~surpassed the best baseline methods by 16.5\% and 21.0\% in NYC, and by 16.1\% and 17.7\% in SEA for Macro-F1 and Micro-F1 scores, respectively. The baseline methods, not designed for different time intervals, showed clear weaknesses, highlighting \MODEL's adaptability to different intervals.

\noindent
\textbf{Changing time granularities of baseline models.}
We trained each baseline model on datasets divided into different interval sizes (6, 12, 24, and 48 hours) and made predictions on test sets with corresponding  interval sizes. In contrast, we trained \MODEL~once and applied it to all time granularities, comparing its performance against the baseline models. 

Figure~\ref{tab:individual category, training on different granularity} shows the prediction results for each crime type dataset. \MODEL~performed well across both common and rare crime datasets, particularly achieving the best performance at a 24-hour granularity and consistently ranking in the top three at 48 hours. While it didn't always secure the top ranking, its flexibility provided significant advantages. Unlike other models that required training and prediction on specific interval lengths, \MODEL~adapted more effectively to varying granularities.

\begin{figure*}[!t]
  \centering
  \begin{minipage}[t]{0.8\linewidth}
    \centering
    \includegraphics[width=\linewidth]{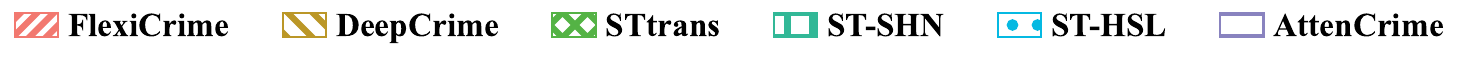}
  \end{minipage}%
  \\
  \centering
  \subfigure[NYC (TG=6 hours)]{
    \begin{minipage}[t]{0.24\linewidth}
      \centering
      \includegraphics[width=\linewidth]{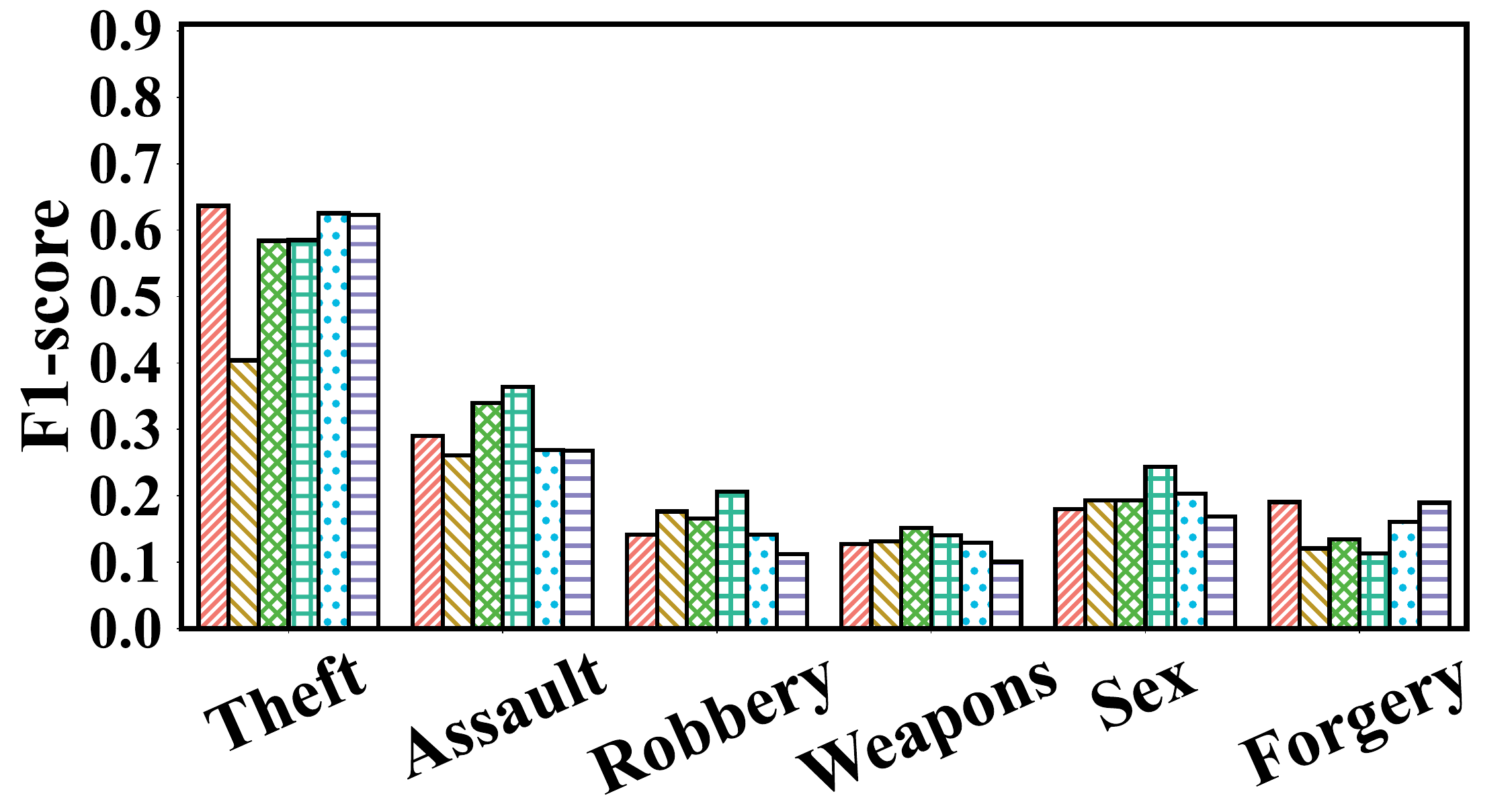}
    \end{minipage}%
  }%
  \subfigure[NYC (TG=12 hours)]{
    \begin{minipage}[t]{0.24\linewidth}
      \centering
      \includegraphics[width=\linewidth]{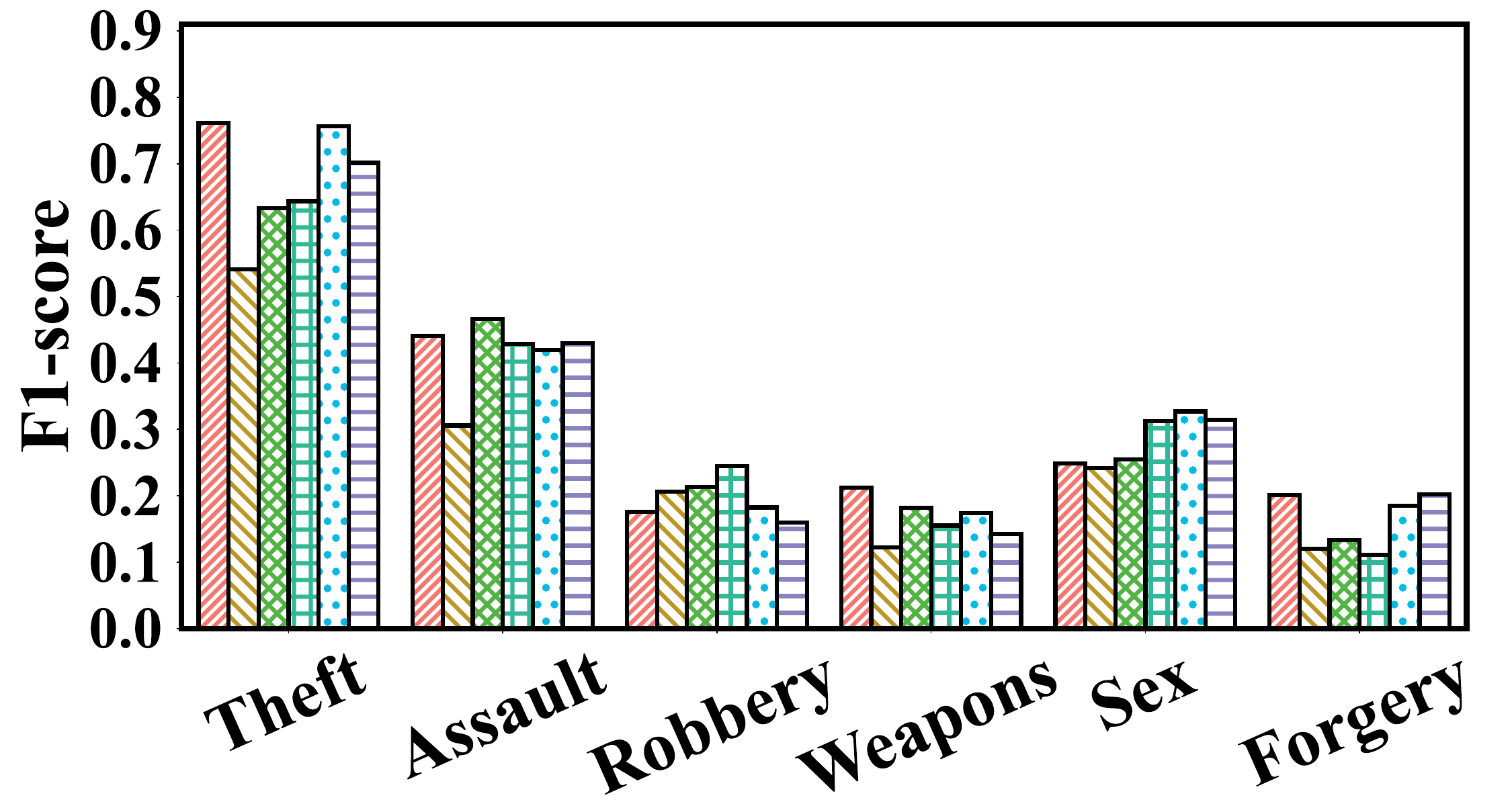}
    \end{minipage}%
  }
  \subfigure[NYC (TG=24 hours)]{
    \begin{minipage}[t]{0.24\linewidth}
      \centering
      \includegraphics[width=\linewidth]{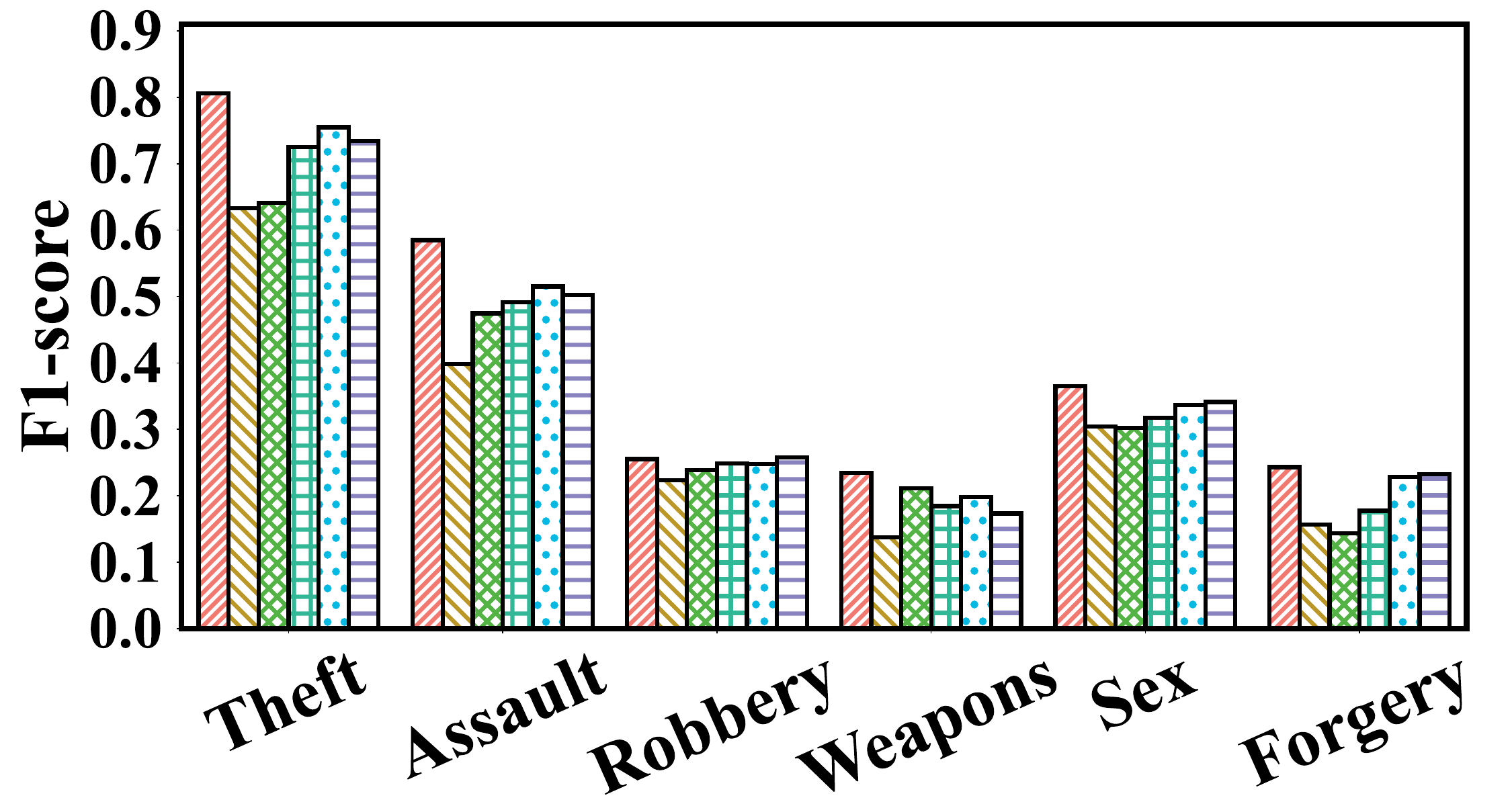}
    \end{minipage}%
  }%
  \subfigure[NYC (TG=48 hours)]{
    \begin{minipage}[t]{0.24\linewidth}
      \centering
      \includegraphics[width=\linewidth]{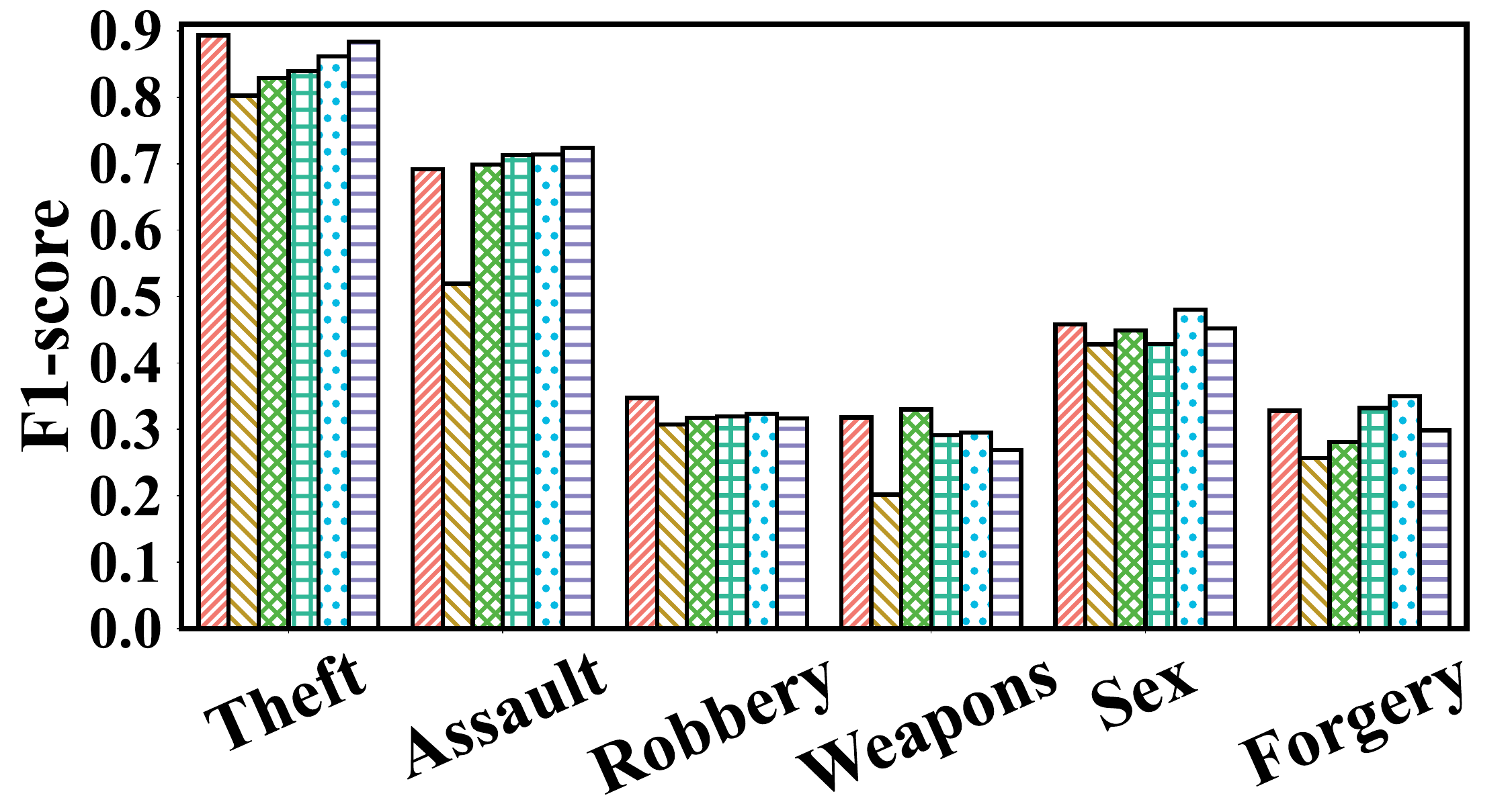}
    \end{minipage}%a
  }\\
  \subfigure[SEA (TG=6 hours)]{
    \begin{minipage}[t]{0.24\linewidth}
      \centering
      \includegraphics[width=\linewidth]{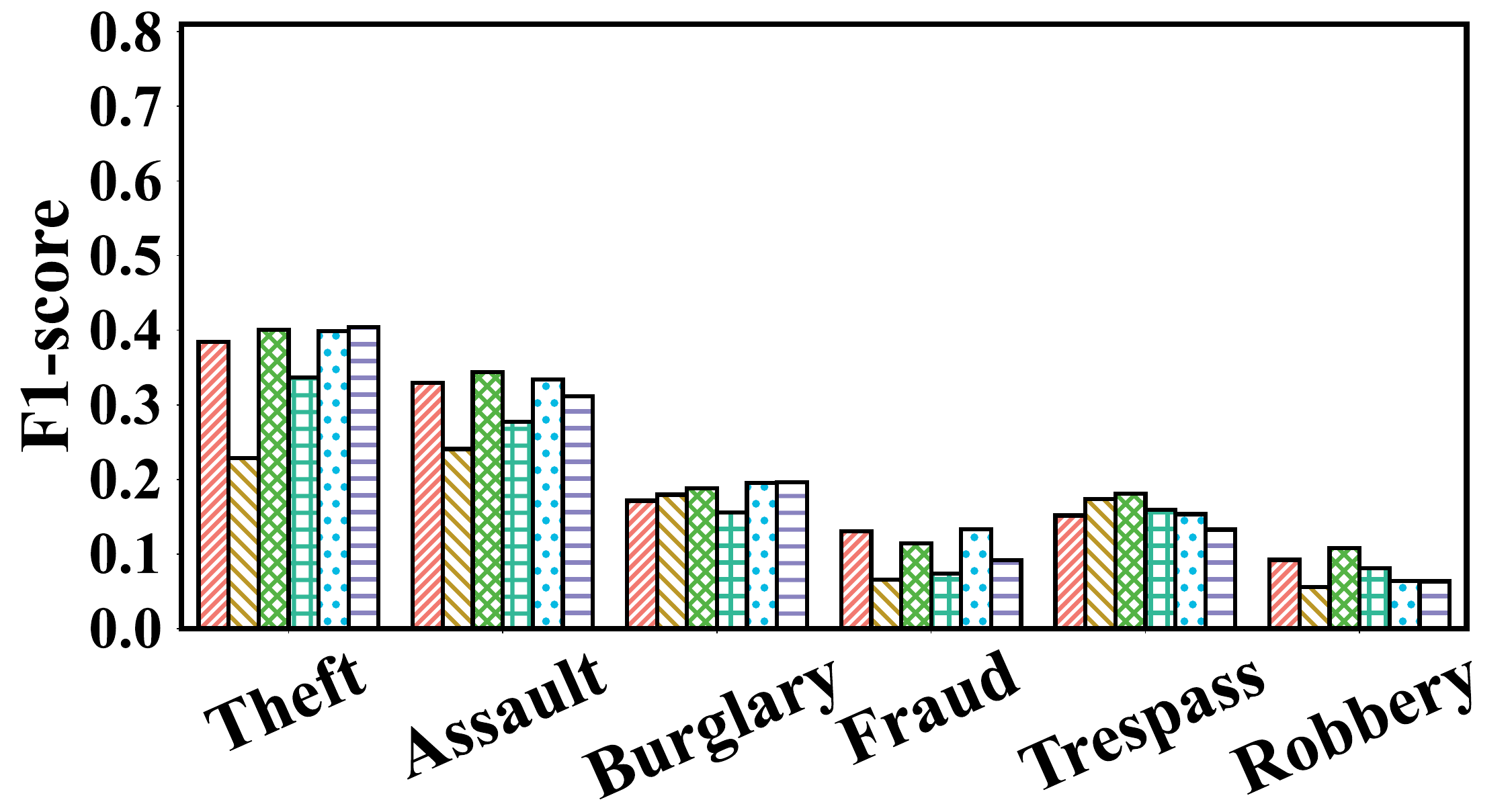}
    \end{minipage}%
  }%
  \subfigure[SEA  (TG=12 hours)]{
    \begin{minipage}[t]{0.24\linewidth}
      \centering
      \includegraphics[width=\linewidth]{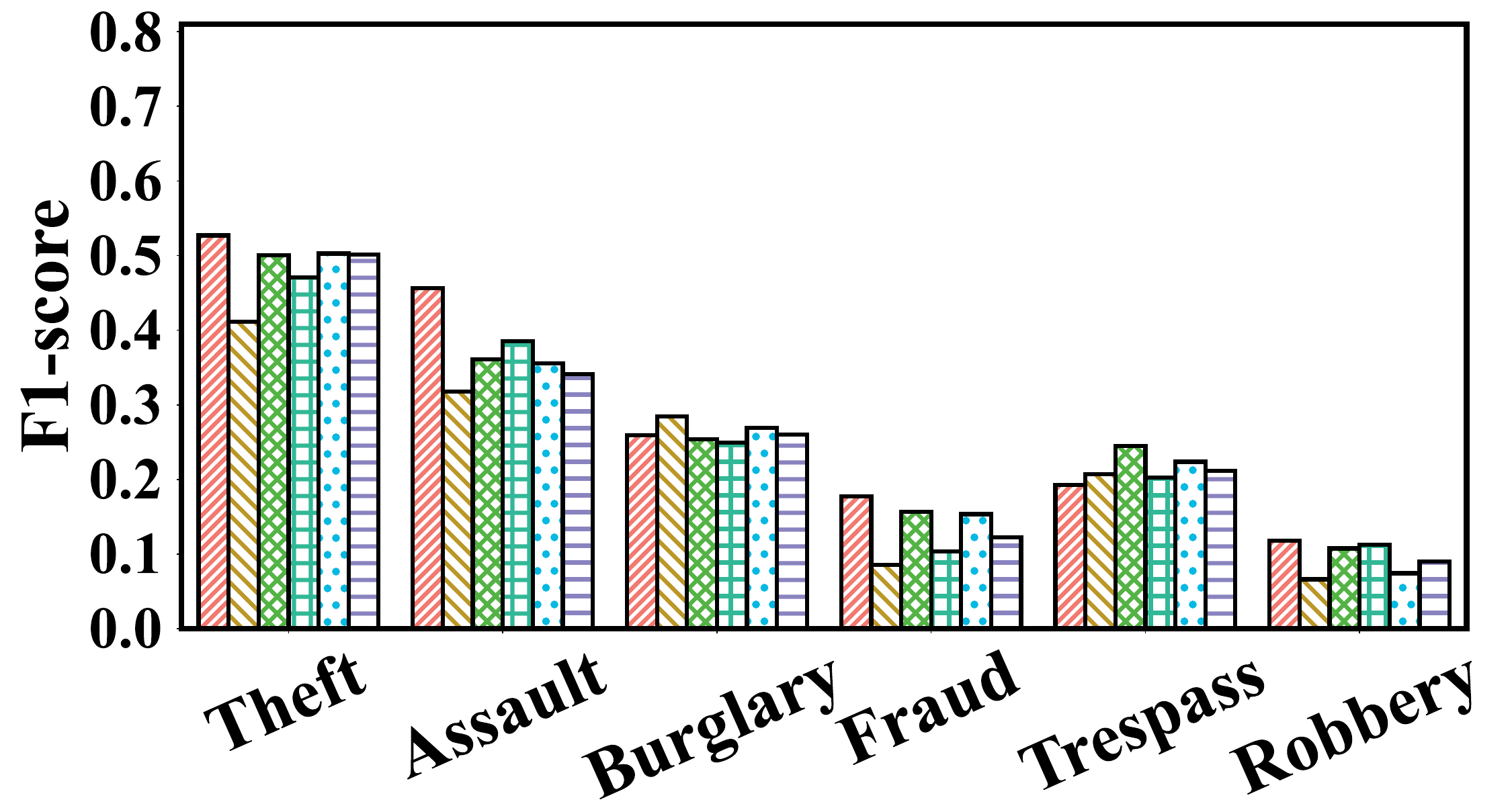}
    \end{minipage}%
  }
  \subfigure[SEA (TG=24 hours)]{
    \begin{minipage}[t]{0.24\linewidth}
      \centering
      \includegraphics[width=\linewidth]{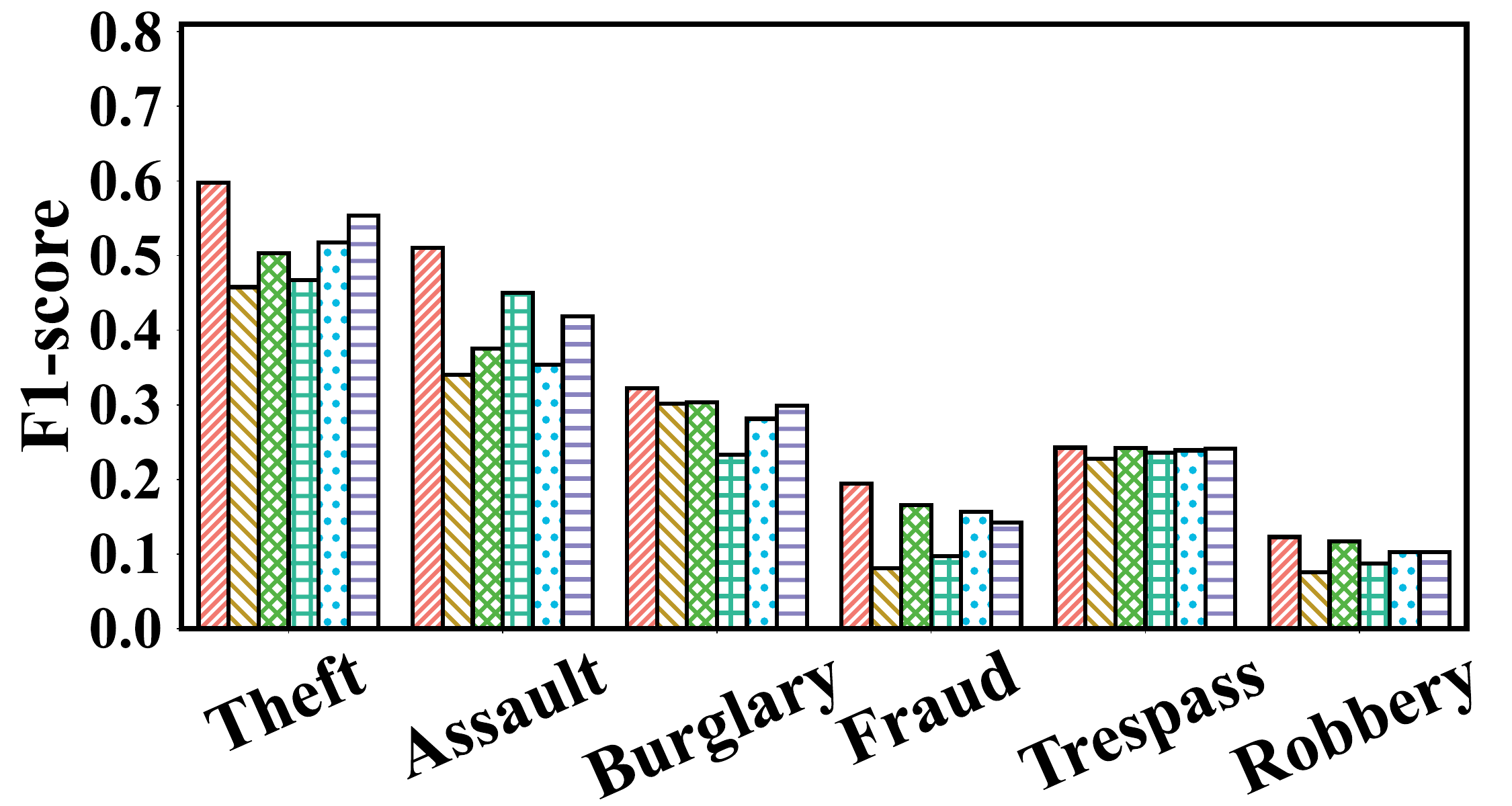}
    \end{minipage}%
  }%
  \subfigure[SEA (TG=48 hours)]{
    \begin{minipage}[t]{0.24\linewidth}
      \centering
      \includegraphics[width=\linewidth]{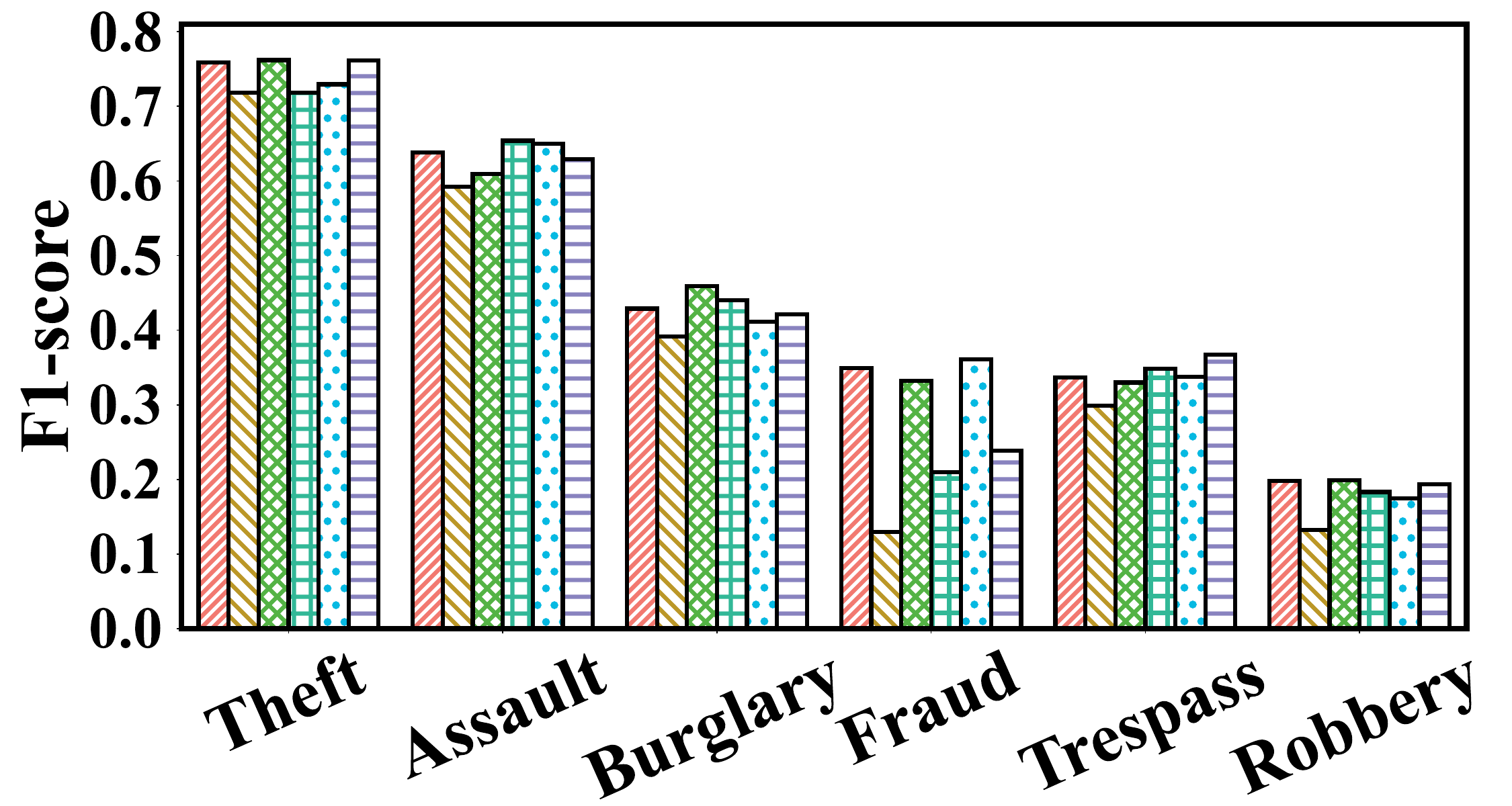}
    \end{minipage}%a
  }%
  \caption{Compare \MODEL~with baseline models across different types of crimes and various time granularities (RQ2). \MODEL~was trained only once, while multiple versions of the baseline models were trained for each time granularity. }
  \label{tab:individual category, training on different granularity}
\end{figure*}

\begin{figure}[t]
  \centering
  \begin{minipage}[t]{\linewidth}
    \centering
    \includegraphics[width=0.8\linewidth]{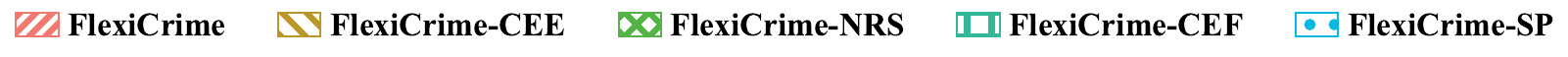}
  \end{minipage}%
  \\
  \subfigure[Common crimes]{
    \begin{minipage}[t]{0.24\linewidth}
      \centering
      \includegraphics[width=\linewidth]{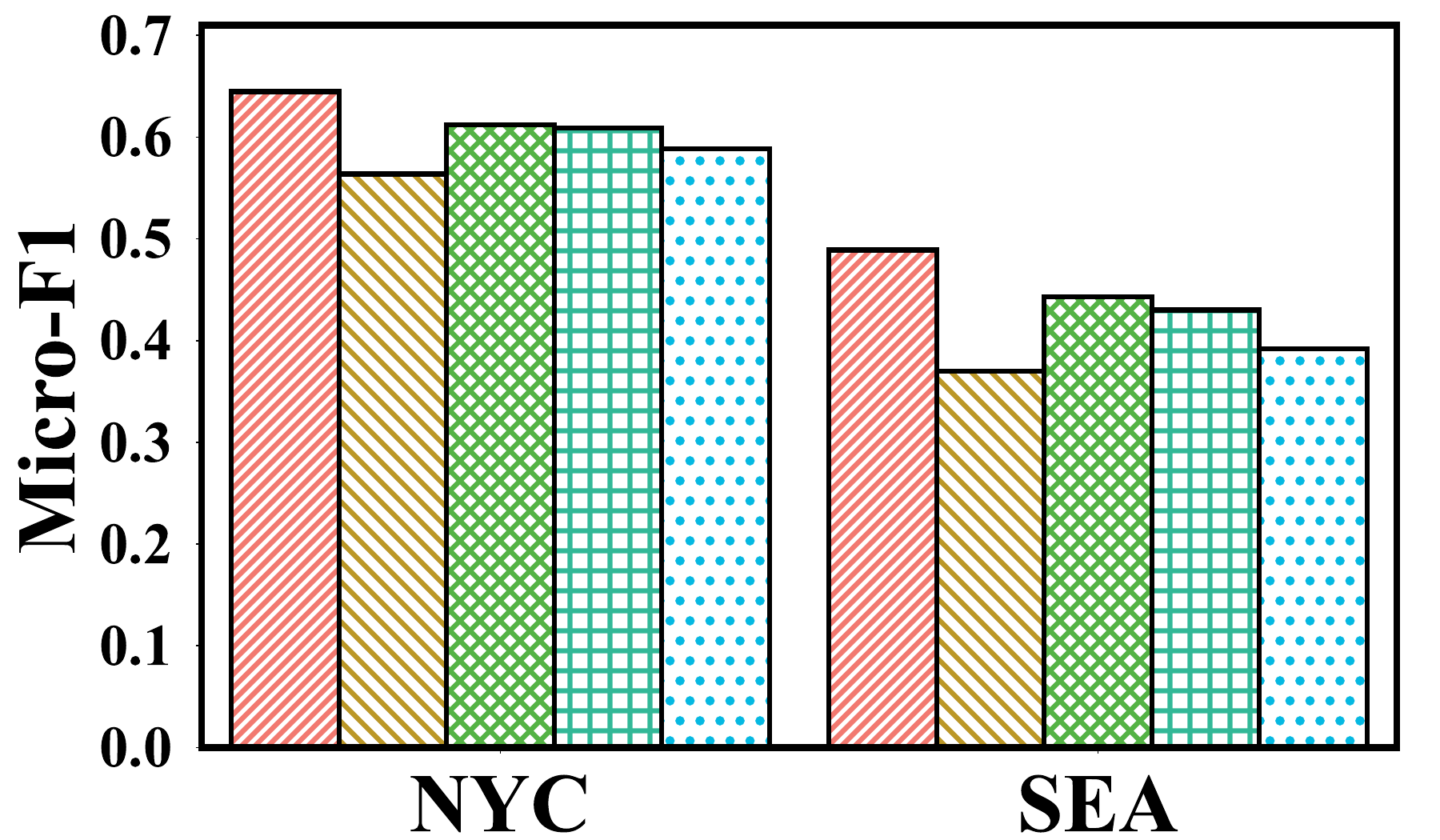}
    \end{minipage}%
    
    \begin{minipage}[t]{0.24\linewidth}
      \centering
      \includegraphics[width=\linewidth]{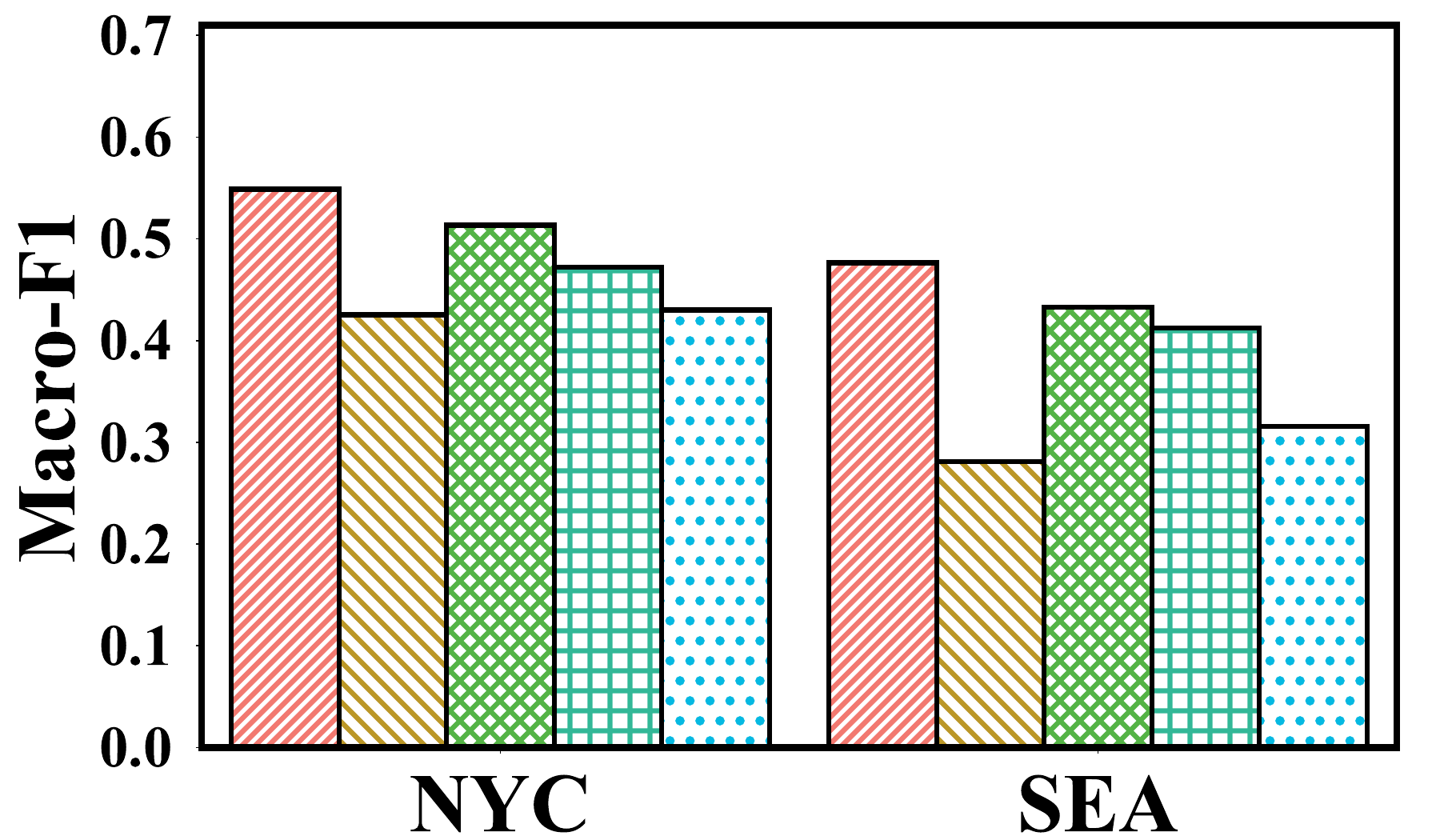}
    \end{minipage}%
  }%
  \subfigure[Rare crimes]{
    \begin{minipage}[t]{0.24\linewidth}
      \centering
      \includegraphics[width=\linewidth]{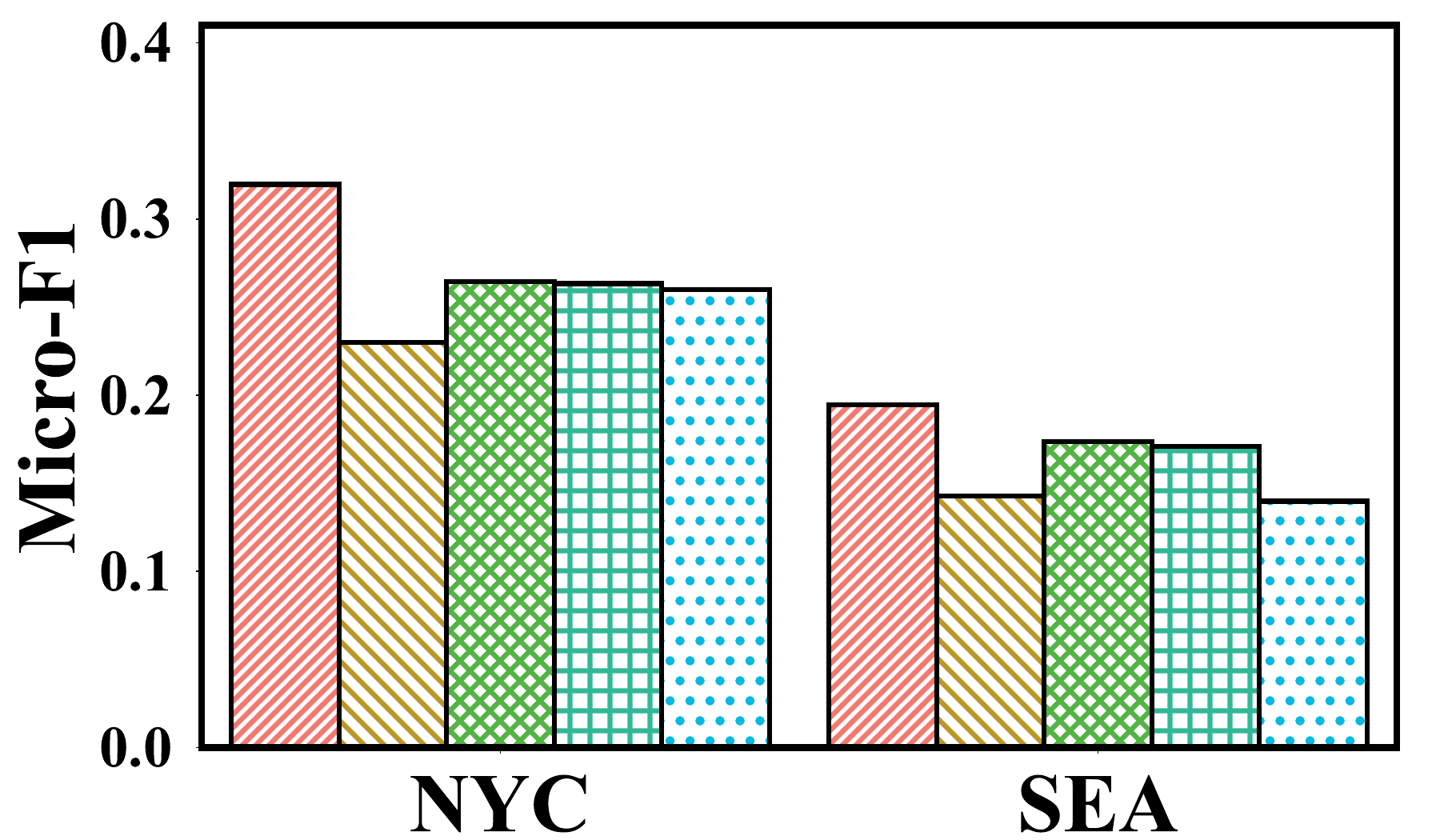}
    \end{minipage}%
    
    \begin{minipage}[t]{0.24\linewidth}
      \centering
      \includegraphics[width=\linewidth]{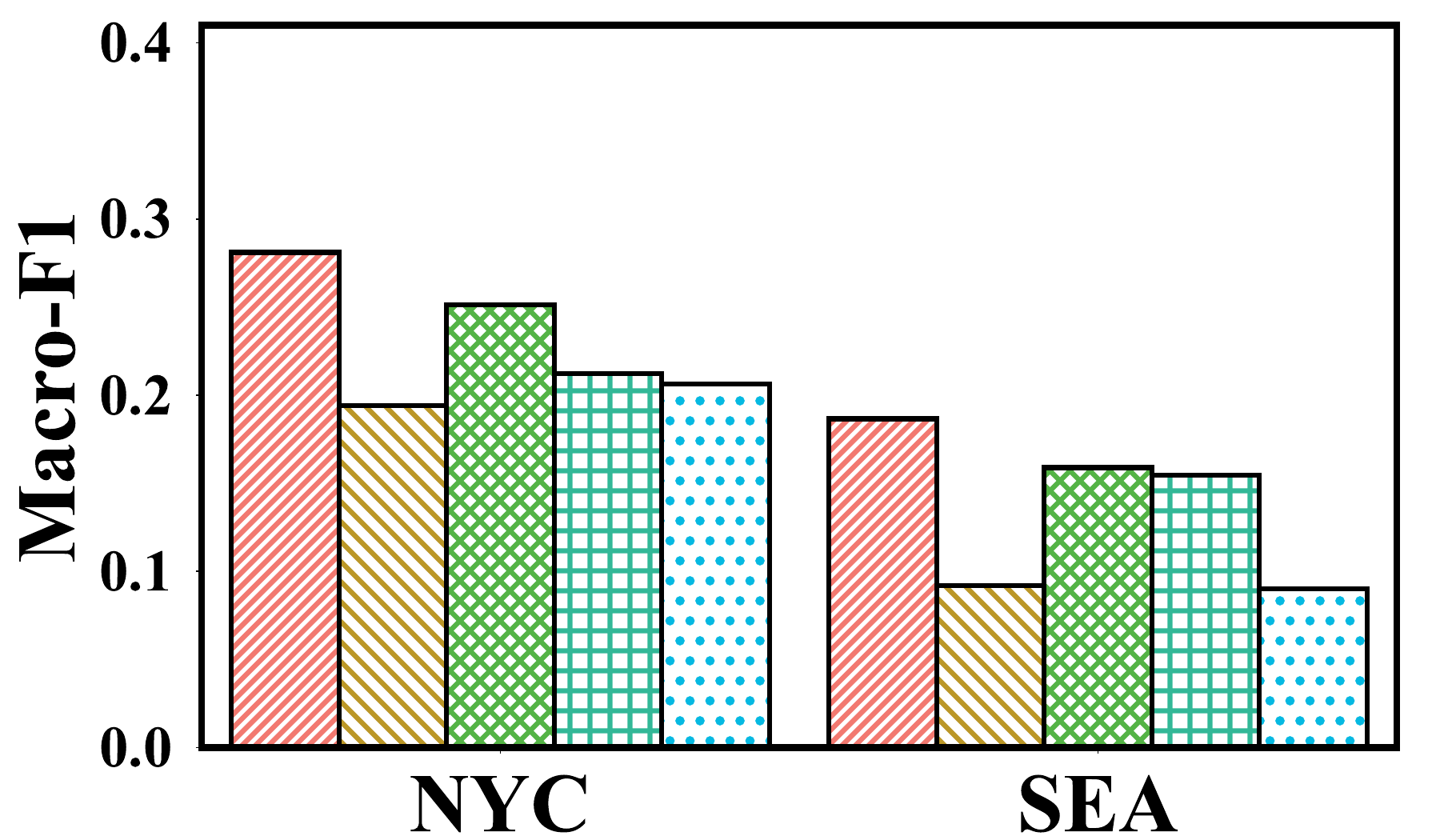}
    \end{minipage}%
  }%
  \centering
  \caption{Ablation experiment result~(RQ3)}
  \label{tab:Ablation Experiment Result}%
\end{figure}

\subsection{Ablation Study (RQ3)}
\label{Sec:rq4}
We evaluated the core modules in \MODEL~by testing four variations: 
(i) ''-CEE'': Removed the crime event encoding component, using MLP encoding instead. 
(ii) ''-NRS'': Excluded near-repeat sampling based on time and location. 
(iii) ''-CEF'': Omitted the crime evolving feature from the type-aware spatiotemporal point process. 
(iv) ''-SP'': Removed sample points, using only the features from the start time of the interval for predictions. 
Results in Figure~\ref{tab:Ablation Experiment Result} show that removing any module degrades prediction performance.

\noindent
{\bf Continuous-time attention network.} (i) ''-CEE'': The crime event encoding module outperforms simple MLP encoding in capturing the attention of time and location of crime events. It increased the Micro-F1 score by 0.1 to 0.15. The target-aware crime event encoder improved \MODEL's ability to extract important features related to crime events. (ii) ''-NRS'': Near-repeat sampling in the continuous-time attention network enhances the model's perception of spatio-temporal correlations between events. It directs the model's attention towards highly correlated crime incidents, improving model robustness. This feature increased the Micro-F1 score by about 0.05 in experiments. 

\noindent
{\bf Type-aware spatiotemporal point process.}
(iii) ''-CEF'': The type-aware spatiotemporal point process model captures crime risk intensities among sampled points, positively impacting prediction accuracy. It increased the Micro-F1 score by 0.05 to 0.1 in experiments. 

\noindent
{\bf Flexible interval prediction.}
(iv) ''-SP'': \MODEL~incorporates sample points to obtain crime context and evolving features during the time interval. Our experiments show that this method increased the Micro-F1 score by 0.05 to 0.2. Using the sample points proves more effective than simply relying on the crime features of the start time point.

\begin{table}[t]
  \setlength{\tabcolsep}{4pt}
  \small
  \centering
  \renewcommand\arraystretch{1}
  \caption{Top-$k$ crime hotspots in terms of HR@$10$~(RQ4)}
  \resizebox{0.75\linewidth}{!}{
    \begin{tabular}{c|c|c|c|c}
      \toprule
      \multirow{2}[4]{*}[1.2ex]{Model} & \multicolumn{2}{c|}{NYC} & \multicolumn{2}{c}{SEA}\\
      \cline{2-5} & \multicolumn{1}{c|}{Common Crimes} & \multicolumn{1}{c|}{Rare Crimes} & \multicolumn{1}{c|}{Common Crimes} & \multicolumn{1}{c}{Rare Crimes} \\
      \cline{1-5}
      DeepCrime   & 0.2352                             & 0.1994                           & 0.2145                             & 0.3319                          \\
      
      STtrans     & \ul{0.3425}                        & \ul{0.5918}                      & \ul{0.4151}                        & 0.5661                          \\
      
      ST-SHN      & 0.3128                             & 0.5574                           & 0.4069                             & \ul{0.5756}                     \\
      
      ST-HSL      & 0.3185                             & 0.4449                           & 0.3874                             & 0.3617                          \\
      
      \midrule
      FlexiCrime  & \textbf{0.6779}                    & \textbf{0.5986}                  & \textbf{0.4427}                    & \textbf{0.7395}                 \\
      \bottomrule
    \end{tabular}%
  }
  \label{tab:HR performance}%
\end{table}%

\subsection{Top-k Crime Hotspots (RQ4)}
We evaluated the alignment of predicted crime hotspots with true hotspots using the Hit Ratio@$k$ (HR@$k$) metric for accuracy. For a target time interval $I$, we computed the probability $\mathbf{X}^{I}$ for each grid and selected the top $k$ grids with the highest probabilities for testing. True hotspots were identified as the top $k$ grids in the ground truth, ranked by crime numbers. Thus, HR@$k$ reflects the proportion of true hotspots among the top-$k$ predicted ones. In this experiment, we set $k$ to 10, trained baseline models with 6-hour granularity, and made predictions at 24-hour intervals.

We conducted experiments on NYC and SEA datasets for common and rare crimes, performing 100 tests for each setting. Table~\ref{tab:HR performance} presents the average HR@$k$ results, showing that \MODEL's HR@$k$ scores were 33.55\% higher than the second-best model. 
These high scores suggest that \MODEL's predicted crime hotspots are likely true hotspots, aiding law enforcement in resource allocation. In contrast, other models struggled with accuracy due to misalignment between training and target time intervals. Overall, our model demonstrated strong adaptability to the changes in time granularity, achieving effective predictions for urban crime hotspots.

\begin{figure*}[t]
  \centering
  \subfigure[Grountruth with various time intervals]{
    \includegraphics[width=0.48\linewidth]{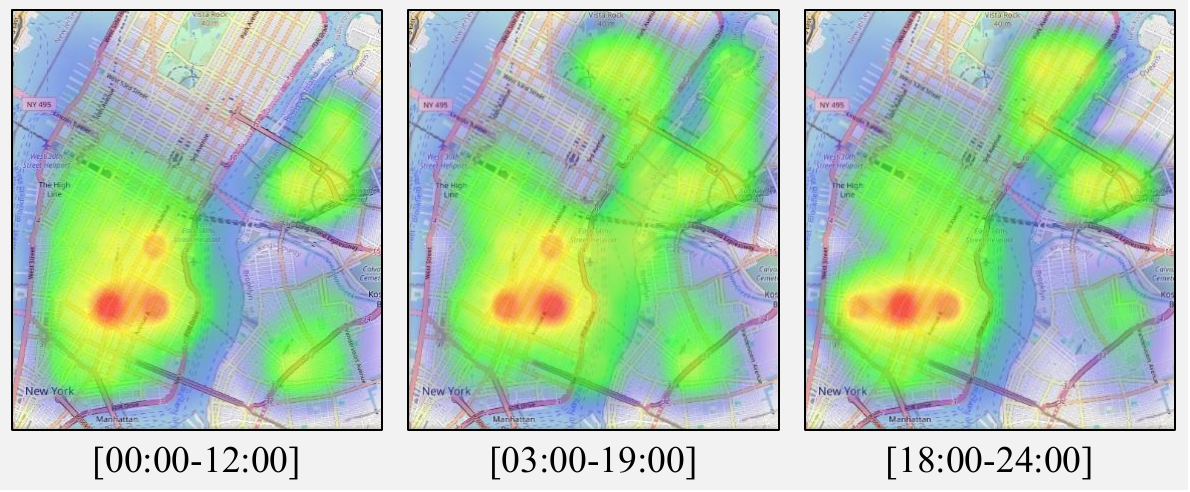}
    \label{fig:Case Study Groundtruth}
  }
  \subfigure[Prediction result of \MODEL]{
    \includegraphics[width=0.48\linewidth]{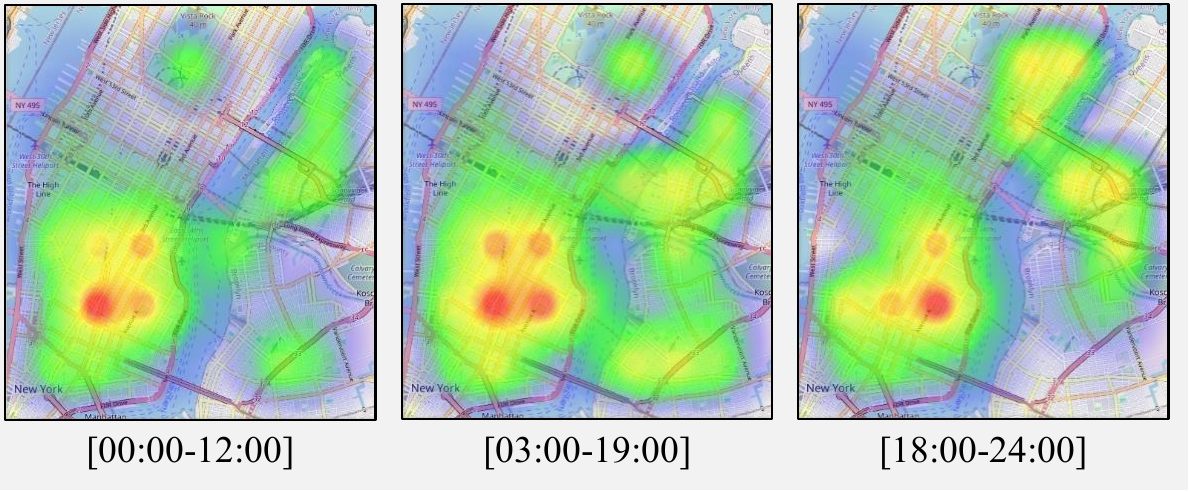}
    \label{fig:Case Study FlexiCrime}
  }
  \\
  \centering
  \subfigure[Groundtruth and prediction results for time interval during 00:00 and 24:00]{
    \includegraphics[width=0.985\linewidth]{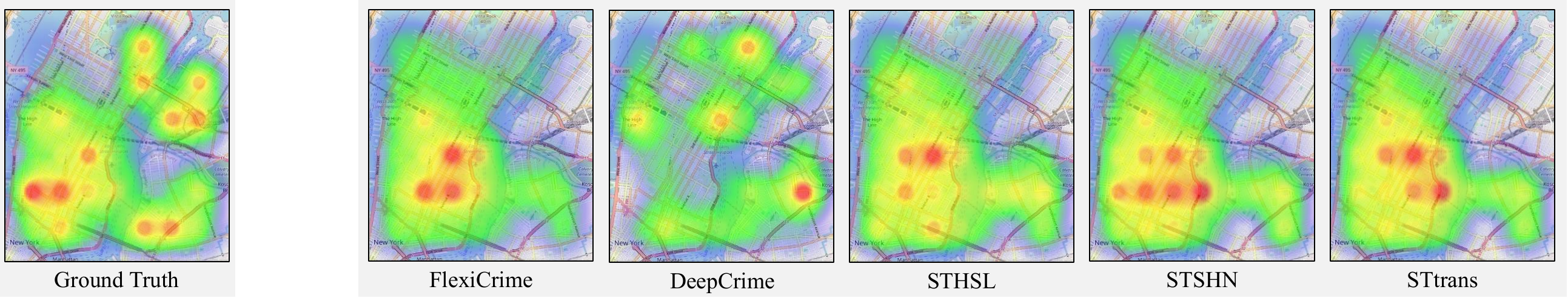}
    \label{fig:Case Study Baseline}
  }
  \caption{Visualizations of Crime Hotspots Prediction in New York on 2018/02/07 (RQ5)}
  \label{fig:Case Study}
\end{figure*}

\subsection{Case Study (RQ5)}
We visualized the results of the prediction of urban crime hotspots in the form of heat maps. We conducted experiments on the NYC dataset, selecting three time intervals—[00:00-12:00], [03:00-19:00], and [18:00-24:00]—on February 7, 2018, to create the tests.
Figures~\ref{fig:Case Study Groundtruth} and \ref{fig:Case Study FlexiCrime} show that the crime hotspots predicted by \MODEL~match most true hotspots. 

We also compared \MODEL~with baseline models during the time interval [18:00-24:00], where baseline models were trained and predicted with a one-day granularity. Figure~\ref{fig:Case Study Baseline} shows that \MODEL~performed similarly or even better in predicting overall crime hotspots, indicating that its predictions identified actual high-risk areas. In contrast, the baseline models' predictions were restricted to the one-day interval. A comparison of Figure~\ref{fig:Case Study Baseline} with Figures~\ref{fig:Case Study Groundtruth} revealed that the daily hotspots predicted by baseline models differed from true hotspots in the intervals [00:00-12:00], [03:00-19:00], and [18:00-24:00]. While baseline models were limited by fixed granularity, \MODEL~offered flexible crime hotspot predictions.

\subsection{Hyperparameter Study (RQ6)}
\label{Sec:rq5}
\begin{figure}[t]
  \centering
  \subfigure{
    \includegraphics[width=0.9\linewidth]{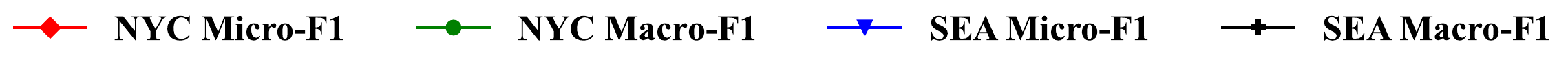}
  }
  \\
  \subfigure{
    \begin{minipage}[t]{0.32\linewidth}
      \centering
      \includegraphics[width=\linewidth]{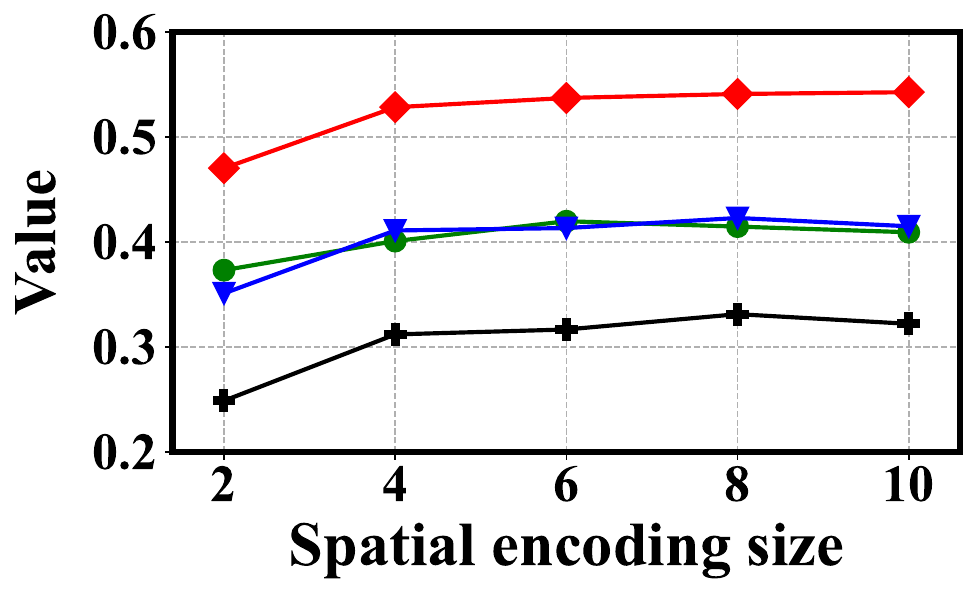}
    \end{minipage}%
  }%
  \subfigure{
    \begin{minipage}[t]{0.32\linewidth}
      \centering
      \includegraphics[width=\linewidth]{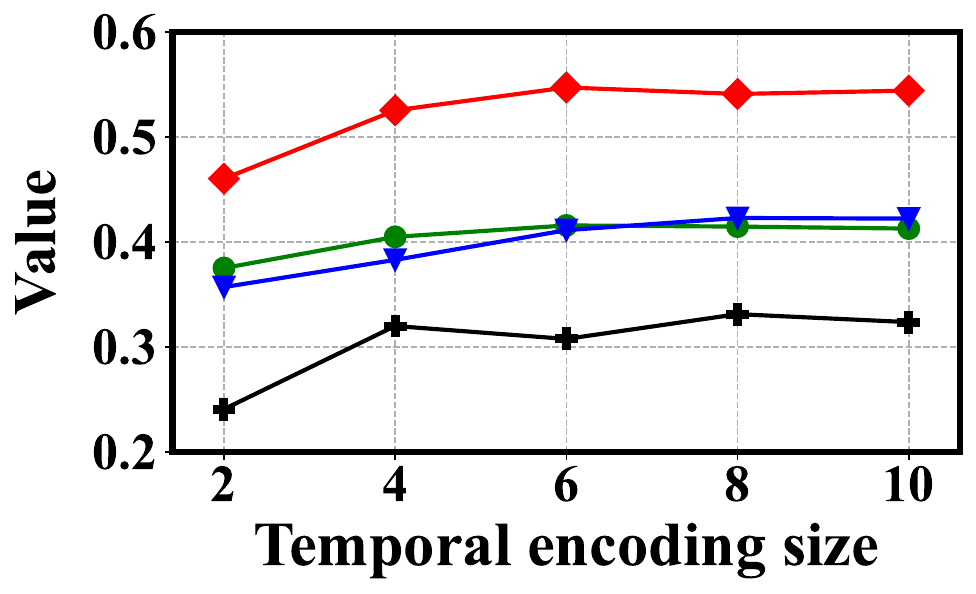}
    \end{minipage}%
  }%
  \subfigure{
    \begin{minipage}[t]{0.32\linewidth}
      \centering
      \includegraphics[width=\linewidth]{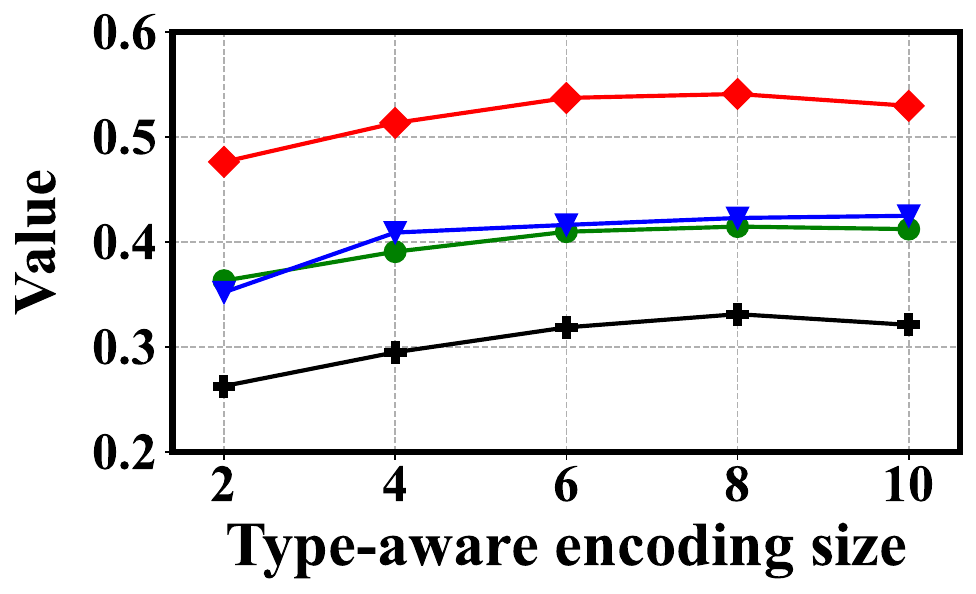}
    \end{minipage}%
  }%
  \\
  \subfigure{
    \begin{minipage}[t]{0.32\linewidth}
      \centering
      \includegraphics[width=\linewidth]{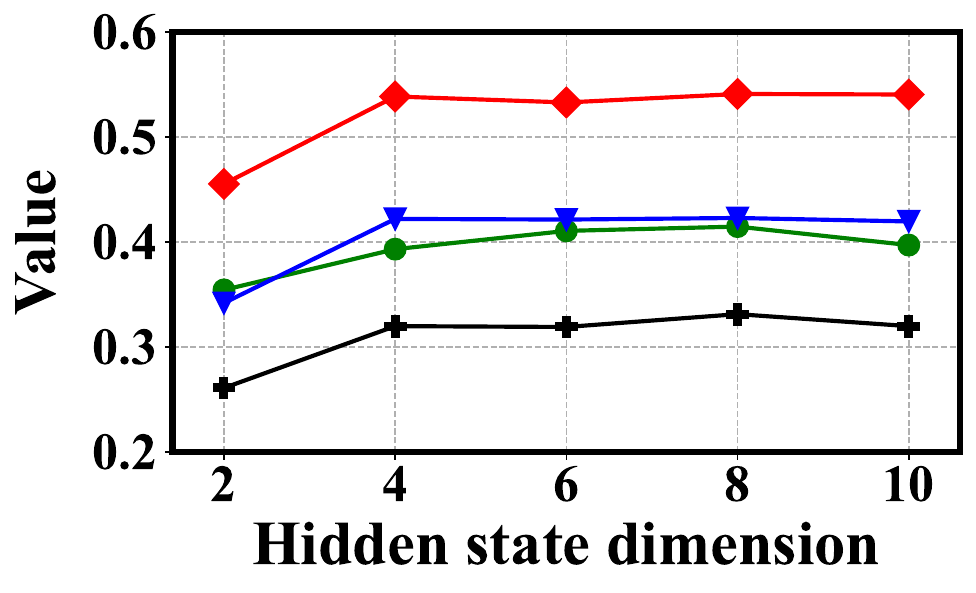}
    \end{minipage}%
  }%
  \subfigure{
    \begin{minipage}[t]{0.32\linewidth}
      \centering
      \includegraphics[width=\linewidth]{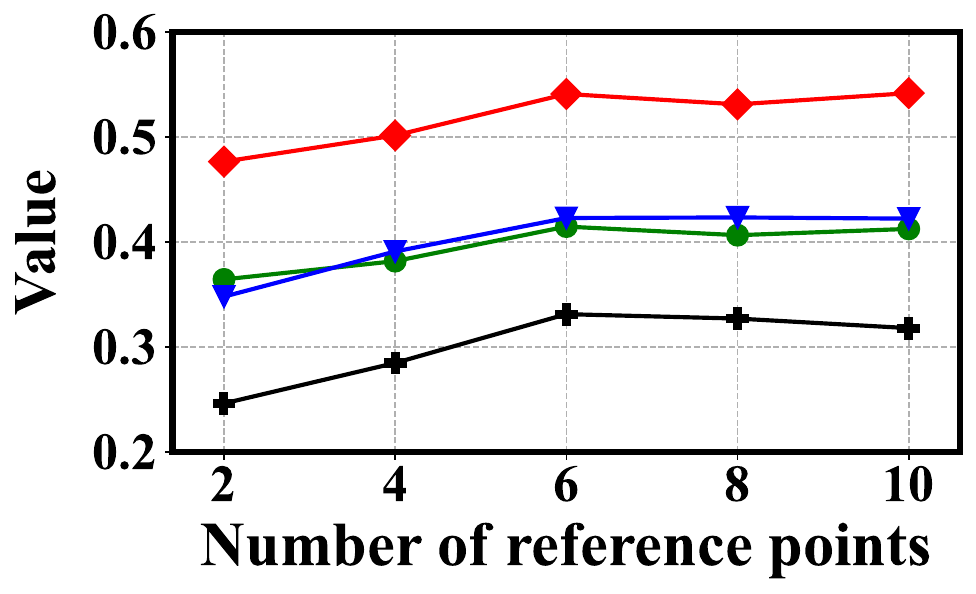}
    \end{minipage}%
  }%
  \subfigure{
    \begin{minipage}[t]{0.32\linewidth}
      \centering
      \includegraphics[width=\linewidth]{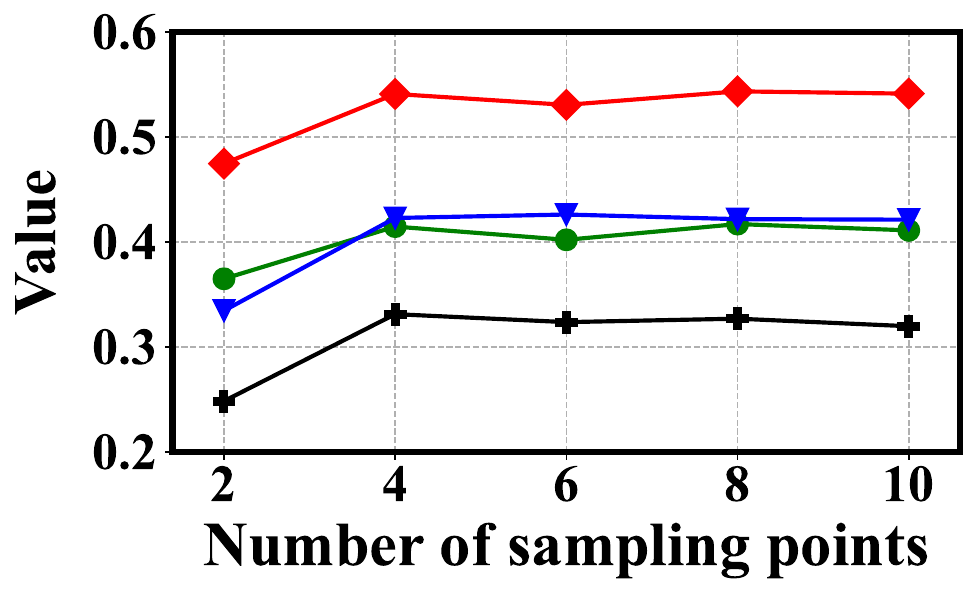}
    \end{minipage}%
  }%
  \centering
  \caption{Hyperparameter study of FlexiCrime~(RQ6)}
  \label{tab:Hyperparameter}%
\end{figure}

We analyzed \MODEL's sensitivity to hyperparameters by examining how different hyperparameter configurations impacted its crime prediction performance. Our focus was on six key parameters: spatial encoding size, temporal encoding size, target-aware event encoding size, the hidden state dimension $\boldsymbol{h}_\tau$, and the number of reference and sample points for flexible interval prediction, while keeping all other parameters at their default values. Figure \ref{tab:Hyperparameter} shows the evaluation results for various parameter choices, measured by Macro-F1 and Micro-F1 metrics.

We found that spatial, temporal, and target-aware event encoding sizes, as well as the dimension of the hidden state $\boldsymbol{h}_\tau$, had minimal impact on \MODEL's performance when exceeding 32. As these sizes increased from 32 to 64, Micro-F1 and Macro-F1 scores changed by less than 0.03 on both datasets. However, performance slightly decreased at sizes of 80, likely due to increased complexity and overfitting, which affected \MODEL's ability of generalization. Setting the encoding sizes to 64 adequately captures the crime context and evolving features at each time point.

We also evaluated the impact of reference and sample points on \MODEL's performance. \MODEL~used the crime context features of reference points to predict those of sample points, achieving the best accuracy with 32 reference points; more may introduce noise during spatiotemporal prediction. We found that 4 sample points were adequate to capture crime features for a one-day interval, with additional sample points yielding only a slight improvement in prediction accuracy.

\section{Conclusion}
\label{sec:conclusion}
In this paper, we proposed \MODEL~for flexible crime predictions, utilizing continuous-time crime contexts and evolving features by applying a sampling operation on flexible intervals rather than fixed time divisions. \MODEL~employs an event-centric spatiotemporal attention network to learn the continuous-time crime context from crime events and a sequence model to capture crime context at specific sampled time points. Moreover, we specifically designed a type-aware spatiotemporal point process to calculate continuous-time evolving features between two sampled time points. Extensive experiments on real-world datasets demonstrate that our model significantly outperforms state-of-the-art methods across various settings.

\bibliographystyle{unsrt}
\bibliography{reference}

\end{document}